\documentclass[10pt,twocolumn,letterpaper]{article}
\usepackage[pagenumbers]{wacv}

\usepackage[accsupp]{axessibility}
\usepackage{microtype}
\usepackage{mathtools}
\usepackage{bm}
\usepackage{multirow}
\usepackage{makecell}
\usepackage{tabularx}
\usepackage{array}
\usepackage{adjustbox}

\usepackage{algorithm}
\usepackage{algpseudocode}

\newcolumntype{Y}{>{\centering\arraybackslash}X}

\newcommand{\brats}{BraTS~2021\xspace}
\newcommand{\mamamia}{MAMA-MIA\xspace}

\usepackage{float}
\usepackage{verbatim}

\usepackage{placeins}
\usepackage{stfloats}

\definecolor{wacvblue}{rgb}{0.21,0.49,0.74}
\usepackage[breaklinks,colorlinks,allcolors=wacvblue]{hyperref}

\hypersetup{%
  pdftitle={TAVO: Target-Aware Source Curation for Budgeted Cross-Center Tumor Segmentation},
  pdfauthor={Xiwen Chen, Tong Miao, Yueqi Dong, Divya Rajparia, Sai Praneeth Karimireddy, Ruishan Liu}%
}

\title{TAVO: Target-Aware Source Curation for\\
Budgeted Cross-Center Tumor Segmentation}

\author{%
Xiwen Chen$^{1}$\thanks{Equal contributions. $^{\dagger}$Corresponding author.}\quad
Tong Miao$^{1}$\footnotemark[1]\quad
Yueqi Dong$^{1}$\\
Divya Rajparia$^{1, 2}$\quad
Sai Praneeth Karimireddy$^{1}$\quad
Ruishan Liu$^{1}$$^{\dagger}$\\
{\normalsize $^{1}$University of Southern California, USA\quad $^{2}$IIT Hyderabad, India}\\
{\small\ttfamily xiwenc@usc.edu, miaot@usc.edu, yueqidon@usc.edu}\\[-1pt]
{\small\ttfamily es22btech11013@iith.ac.in, karimire@usc.edu, ruishanl@usc.edu}\\[6pt]
\parbox{0.93\textwidth}{\normalfont\footnotesize\raggedright
\textcopyright{} 2027 IEEE. Personal use of this material is permitted. Permission from IEEE must be obtained for all other uses, in any current or future media, including reprinting/republishing this material for advertising or promotional purposes, creating new collective works, for resale or redistribution to servers or lists, or reuse of any copyrighted component of this work in other works.
}%
}

\begin{document}

\maketitle

\begin{abstract}
Tumor segmentation at a new clinical center often relies on a small labeled target set and a heterogeneous pool of annotated external cases. Because tumor labels are costly and imaging protocols, patient populations, and tumor appearances vary across centers, using all external cases can increase training cost and induce negative transfer. We introduce \textbf{Target-Aware Valuation Optimization (TAVO)}, a
model-external source-curation framework for budgeted cross-center tumor
segmentation. Given a source-case budget \(B\), TAVO adaptively combines target similarity, gradient
compatibility, source coverage, and diversity by normalizing heterogeneous criterion scores into a common rank space and learning target- and budget-specific simplex weights from held-out target validation utility. The selected top-$B$ cases enter standard segmentation training without changing the architecture, loss, or inference. We evaluate TAVO on BraTS 2021 and MAMA-MIA, two multicenter MRI tumor-segmentation benchmarks covering eight held-out target centers or domains, with OfficeHome as a compact cross-task control. At a 150-case source budget, TAVO improves target-averaged Dice over the strongest fixed selector by 1.0 points on BraTS and 1.9 points on MAMA-MIA, and over the strongest budget-matched adaptation baseline by 1.0 and 2.2 points, respectively. These results show that target- and budget-adaptive source curation can be more reliable than fixed selection rules and can outperform Target + Full Source while using substantially fewer external cases during final training. Code: \href{https://github.com/francischen0724/TAVO}{https://github.com/francischen0724/TAVO}.

\end{abstract}

\begin{figure}[t]
    \centering
    \includegraphics[width=0.75\linewidth]{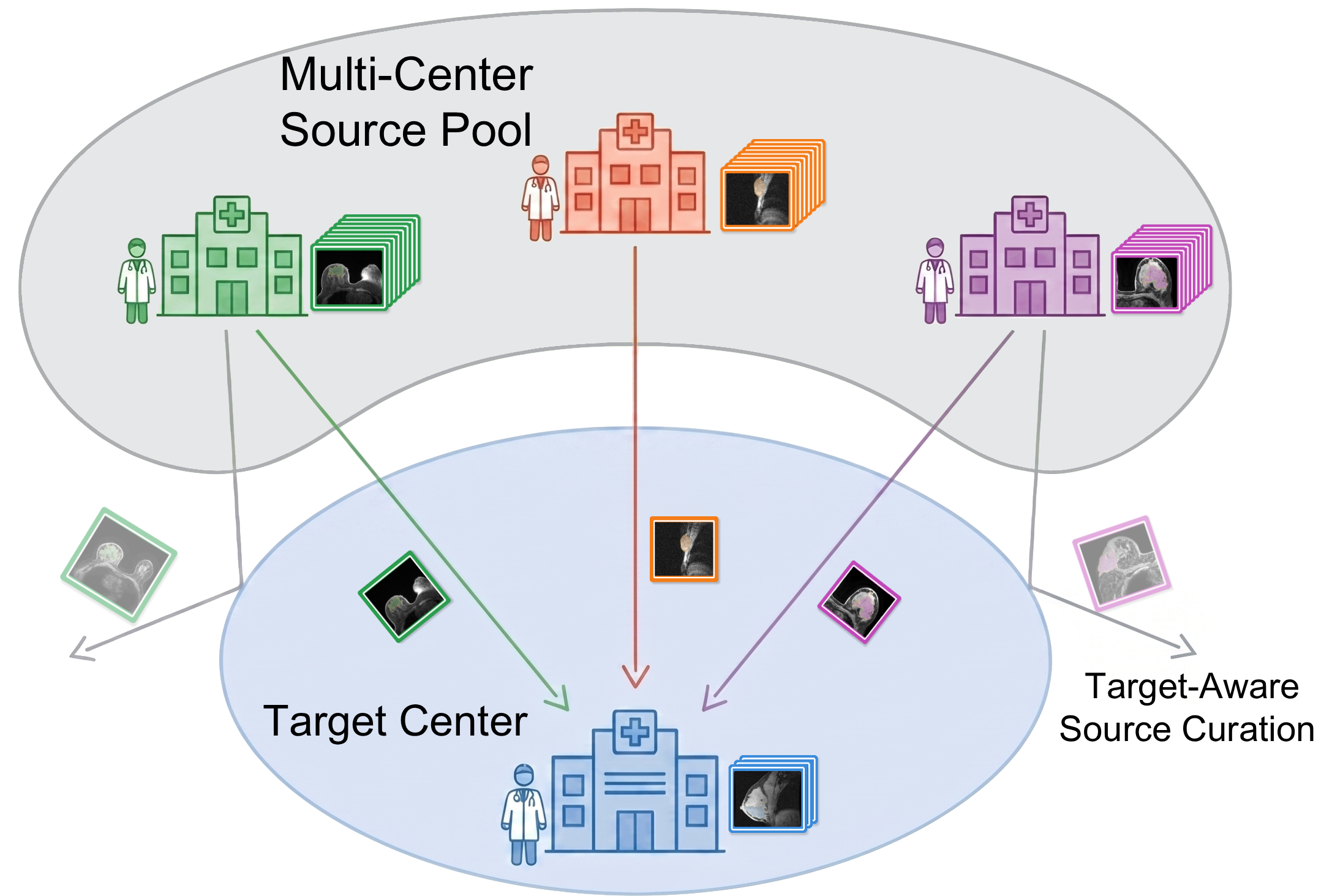}
    \caption{
Cross-center source-curation setting. Given limited target support and
a heterogeneous external source pool, TAVO curates budgeted source
cases before standard segmentation training.
}
    \label{fig:intro_motivation}
\end{figure}
\section{Introduction}
\label{sec:introduction}

Tumor segmentation is a particularly challenging setting for cross-center model development. For a new target center, only a small number of labeled cases may be available, because volumetric tumor delineation requires expert voxel-level annotation and tumor appearance can vary substantially in size, shape, location, contrast, and subtype. Meanwhile, annotated external cases may be available from other institutions, clinical cohorts, or public datasets. However, simply pooling all external cases is not necessarily beneficial: scanners, acquisition protocols, treatment cohorts, and patient populations differ across centers, and poorly matched cases can increase computational cost or induce negative transfer \citep{guan2022domainadapt,peng2021limitedsupervision, yoon2024dgreview,zhang2023negativetransfer,zhang2023selectivity}. Thus, for cross-center tumor segmentation, source composition becomes a first-order design choice: the practical question is not only how to adapt the segmentation model, but also which external cases should be admitted to training for a particular target center. \Cref{fig:intro_motivation} illustrates this cross-center source-curation setting.

Motivated by this setting, we study \emph{budgeted source curation} for
cross-center tumor segmentation. We refer to externally annotated cases from
other centers or cohorts as source cases. Given a heterogeneous labeled source
pool, a small labeled target support set, and a source-case budget \(B\),
the goal is to construct a target-adapted training set that improves
segmentation at the target center. The budget is case-level: selecting one
tumor case counts once toward \(B\) and includes all associated slices or patches. Unlike target acquisition, which labels new target cases, we use a fixed
labeled target support to curate useful external source cases from an already
annotated pool.

We present \textbf{Target-Aware Valuation Optimization (TAVO)}, a
model-external source-curation framework for cross-center tumor segmentation. Given a target center and source-case budget \(B\), TAVO computes a set of complementary case-level source rankings from a shared warm-up model, covering target similarity, gradient compatibility, source coverage, and diversity. It normalizes these heterogeneous criterion scores into a common rank space and learns target- and budget-specific fusion weights using held-out target validation utility. The fused ranking is converted into a deterministic top-\(B\) source-case manifest and consumed by an otherwise standard segmentation training pipeline. TAVO changes neither the segmentation architecture nor the task loss and introduces no inference-time overhead. Its output is a curated training set rather than a new deployment model, making it
orthogonal to model-centric adaptation.

We evaluate TAVO on two multicenter MRI tumor-segmentation benchmarks:
multimodal brain tumor segmentation from \brats and breast tumor segmentation
from \mamamia \citep{baid2021brats,garrucho2025mamamia}. Together, these
experiments cover two anatomies, two MRI protocols, eight held-out target
centers, multiple case-level source budgets, and two segmentation
backbones. We compare TAVO with random selection, eight fixed source-selection
criteria, budget-matched model-centric adaptation baselines, and non-budgeted
full-source training. The results show that the strongest fixed criterion
changes across targets and budgets, while TAVO provides consistently strong
target-averaged performance and can exceed Target + Full Source while using substantially fewer external cases during final training. We retain OfficeHome as a compact
cross-task control to test whether the same rank-fusion interface transfers beyond MRI tumor segmentation \citep{venkateswara2017deep}.

Finally, we summarize our contributions as follows:
\begin{itemize}
    \item We formulate budgeted external-case curation for cross-center tumor
    segmentation, where limited labeled target support guides which externally
    annotated tumor cases are admitted to downstream training.

    \item We introduce TAVO, a model-external data interface that combines
    target similarity, gradient compatibility, source coverage, and diversity
    into a target- and budget-specific source-case manifest while leaving the
    segmentation architecture, task loss, and inference procedure unchanged.

    \item We provide a comparative assessment on two multicenter MRI
    tumor-segmentation benchmarks, covering eight held-out target centers or
    domains, multiple case budgets, eight source-selection criteria,
    budget-matched adaptation baselines, and two segmentation backbones.
\end{itemize}
\section{Related Work}
\label{sec:related_work}

\paragraph{Cross-center tumor segmentation.}
Tumor segmentation benchmarks such as BraTS and MAMA-MIA have enabled
systematic evaluation across institutions, protocols,
and patient cohorts \citep{baid2021brats,garrucho2025mamamia}. Standardized
segmentation pipelines such as nnU-Net provide strong task-specific baselines,
but cross-center variation remains a persistent challenge when a model trained
on pooled external data is applied to a new target center
\citep{isensee2021nnunet,guan2022domainadapt,yoon2024dgreview}. Most work in
this area improves architectures, preprocessing, augmentation, or adaptation
objectives while assuming that the source training set is fixed. Our focus is
orthogonal: we ask which externally annotated tumor cases should be selected
under a case-level source budget before standard downstream training.

\paragraph{Data subset selection and valuation.} Data subset selection seeks compact training sets that preserve useful learning signals while reducing training cost. Submodular objectives and geometric coresets prioritize coverage, representativeness, or diversity \citep{wei2015submodularity,sener2018coreset}, while optimization-aware methods select examples based on training dynamics or gradient approximation. CRAIG approximates the full-data gradient, GradMatch matches subset gradients to a
training or validation objective, and GLISTER optimizes held-out utility through bilevel subset selection \citep{mirzasoleiman2020craig,killamsetty2021gradmatch,killamsetty2021glister}. Related data-valuation methods estimate scalar example contributions using influence functions, Data Shapley, or training-trajectory attribution~\citep{koh2017influence,ghorbani2019datashapley,pruthi2020tracin}. These methods provide useful notions of coverage, optimization relevance, and example value, but they do not by themselves specify how heterogeneous criteria
should be combined for a particular target center and source budget.

\paragraph{Target-aware data selection.}
Several methods use held-out or target data to adapt data importance. Meta-reweighting and Meta-Weight-Net learn sample weights from clean
meta-data \citep{ren2018reweight,shu2019metaweightnet}, while DVRL and DDS learn data-value estimators from downstream
feedback \citep{yoon2020dvrl,wang2020dds}. Under domain shift, ORIENT uses target-query submodular mutual information to balance source--target relevance and diversity \citep{karanam2022orient}; Selectivity prunes source data using source--target label and feature mappings \citep{zhang2023selectivity}; and LESS identifies examples through target-relevant gradient similarity \citep{xia2024less}. These methods show that source utility can depend on target evidence, validation feedback, or an available budget. However, they typically instantiate a single predefined selection objective, a single learned valuation mechanism, or a fixed weighting policy. TAVO instead
optimizes a target- and budget-specific simplex combination of heterogeneous criteria, retaining single-criterion selectors as simplex corners and inducing a hard top-\(B\) source subset.

\paragraph{Model-centric domain adaptation and multi-source transfer.} Domain adaptation addresses source--target shift primarily by modifying the learning objective or representation space. Discrepancy-based methods such as DAN/MMD reduce distribution mismatch with kernel distances \citep{long2015dan}, and CORAL aligns second-order feature statistics \citep{sun2016coral}. Adversarial methods such as DANN and CDAN instead learn domain-invariant representations \citep{ganin2016dann,long2018cdan}. For segmentation, ADVENT performs adversarial entropy minimization over structured outputs, and medical adaptation methods such as SE-ASA combine selective entropy constraints with semantic alignment \citep{vu2019advent,feng2023seasa}. Few-shot or semi-supervised adaptation methods such as MME and CLDA use limited target
labels, often together with unlabeled target data \citep{saito2019mme,singh2021clda}, and multi-source methods such as MDAN and M3SDA model multiple source domains through adversarial bounds or moment matching \citep{zhao2018mdan,peng2019m3sda}. These methods adapt the learner or aggregate domains, whereas TAVO addresses the sample-level source-composition decision under an explicit cardinality budget.
\section{Methods}
\label{sec:method}

\begin{figure*}[t]

    \centering
    \includegraphics[width=\textwidth]{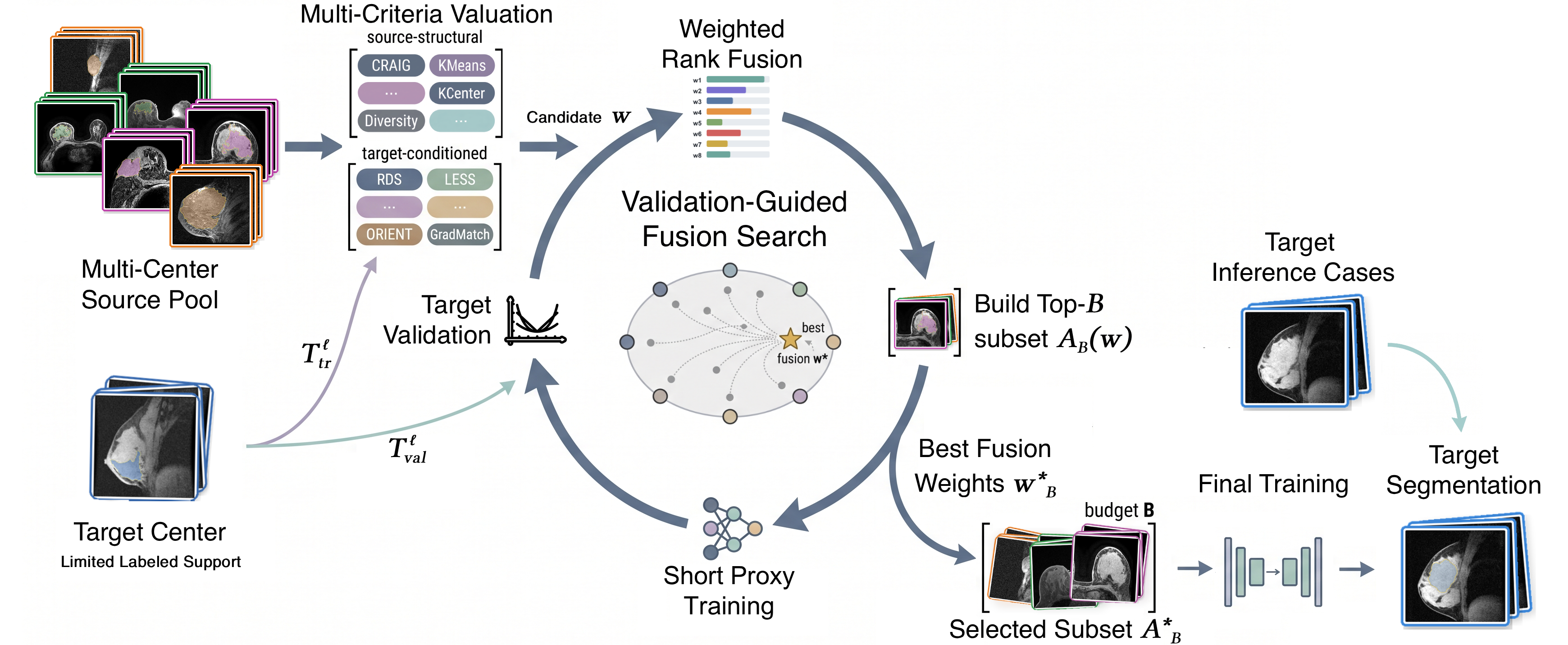}
    \caption{
    Overview of TAVO.
    Given a heterogeneous labeled source pool and limited labeled target support,
    TAVO computes complementary case-level source rankings from target-conditioned
    and source-structural criteria. These rankings are aligned in a common rank
    space and combined by candidate fusion vectors. Each candidate \(w\) induces
    a top-\(B\) source subset \(A_B(w)\), which is assessed through short proxy
    training and held-out target validation. The selected fusion
    \(w_B^\star\) is frozen, and the resulting source-case manifest
    \(A_B^\star\) is used together with the labeled target training support for
    final tumor-segmentation training.
    }
    \label{fig:overview}
   
\end{figure*}

\Cref{fig:overview} gives an overview of TAVO. We first formalize budgeted source curation for cross-center tumor segmentation, then describe how complementary source rankings are aligned and fused, and finally present the validation-guided multi-fidelity search used to select target- and budget-specific fusion weights. After the top-\(B\) source-case manifest is fixed, final training follows the same downstream architecture, task loss, checkpointing rule, and inference procedure as the comparison methods.

\subsection{Problem Setting}
\label{sec:problem_setting}

Let
\(\mathcal{S}=\bigcup_{d=1}^{D}\mathcal{S}_{d}
=\{u_i=(x_i,y_i)\}_{i=1}^{N}\)
denote a labeled source pool collected from \(D\) external domains. In tumor
segmentation, each source example \(u_i\) is a complete patient case or volume;
selecting it includes all associated slices used by the downstream learner. The labeled target data are split into disjoint training, validation, and test sets,
\(\mathcal{T}_{\mathrm{tr}}^\ell\), \(\mathcal{T}_{\mathrm{val}}^\ell\), and
\(\mathcal{T}_{\mathrm{te}}\). The training support provides supervision and
target evidence for target-conditioned criteria. The validation support provides
the outer-loop utility for fusion selection and checkpoint selection, but is
not used to compute criterion scores or train final models. The test set is
reserved for final evaluation.

Let \(\theta_{\mathrm{init}}\) denote the common task-specific model
initialization. For a candidate source subset
\(\mathcal{A}\subseteq\mathcal{S}\), define
\begin{equation}
\theta(\mathcal{A},\xi)
=
\operatorname{Train}_{\mathrm{full}}
\left(
\theta_{\mathrm{init}};
\mathcal{A}\cup\mathcal{T}_{\mathrm{tr}}^\ell,
\xi
\right),
\label{eq:full_training}
\end{equation}
where \(\xi\) represents training randomness. Given a source-cardinality
budget \(B\leq N\), the ideal subset would maximize held-out target validation utility,
\begin{equation}
\mathcal{A}^{\mathrm{opt}}_{B}
\in
\arg\max_{\substack{
    \mathcal{A}\subseteq\mathcal{S}\\
    |\mathcal{A}|=B}}
\;
\mathbb{E}_{\xi}
\left[
Q
\left(
f_{\theta(\mathcal{A},\xi)};
\mathcal{T}_{\mathrm{val}}^\ell
\right)
\right],
\label{eq:ideal_selection}
\end{equation}
where \(Q\) denotes the target-validation metric evaluated on \(\mathcal{T}_{\mathrm{val}}^\ell\). Exhaustively evaluating the \(\binom{N}{B}\) possible subsets is infeasible, so TAVO restricts the
search to a structured family of subsets induced by a low-dimensional fusion of source-valuation criteria. The source-case budget \(B\) constrains the number of source examples used in candidate and final training.

\subsection{Target-Aware Multi-Criteria Rank Fusion}
\label{sec:rank_fusion}

TAVO combines \(M\) source-valuation criteria through a common ranking interface. In our segmentation experiments, we use \(M=8\) complementary criteria organized into two families. The target-conditioned criteria are RDS~\citep{ruder2017select,ma2019domain}, LESS~\citep{xia2024less}, ORIENT~\citep{karanam2022orient}, and GradMatch~\citep{killamsetty2021gradmatch}. The source-structural criteria are CRAIG~\citep{mirzasoleiman2020craig}, KMeans~\citep{lloyd1982least}, KCenter~\citep{sener2018coreset}, and Diversity~\citep{gonzalez1985clustering}. Together, these criteria provide complementary views of target relevance, gradient compatibility, source coverage, and diversity, allowing TAVO to trade off target-conditioned and source-structural evidence rather than relying on one fixed selection rule.

To compute the feature and gradient evidence required by these criteria, we construct a fixed criterion warm-up model \(f_{\theta_{\mathrm{warm}}}\). Starting from \(\theta_{\mathrm{init}}\), the warm-up model is briefly trained on \(\mathcal{S}\cup\mathcal{T}_{\mathrm{tr}}^\ell\) and is then used only to extract representations and gradients for source valuation. Candidate and final models are reinitialized from \(\theta_{\mathrm{init}}\). This full-pool pass is used only for valuation evidence; the budget \(B\) constrains the source cases used in candidate and final training. The warm-up schedule and criterion-specific implementations are provided in Supplementary Secs.~A.1--A.2.

Each criterion produces either a per-example utility score or an ordered prefix of selected source cases. Score-based criteria are sorted by descending utility, while subset-producing criteria retain their native selection order. Examples outside a returned prefix receive a common tail rank below the listed examples. Criteria whose native output depends on the requested subset size are recomputed separately for each budget \(B\).

Because native utilities are not calibrated across criteria, TAVO fuses ranks
rather than raw scores. Let
\(\rho_{i,B}^{(m)}\in[1,N]\) denote the descending rank assigned to source
example \(i\) by criterion \(m\) under budget \(B\), where rank \(1\)
indicates the highest estimated utility. We define
\begin{equation}
v_{i,B}^{(m)}
=
\frac{N-\rho_{i,B}^{(m)}}{N-1},
\qquad
v_{i,B}^{(m)}\in[0,1].
\label{eq:rank_normalization}
\end{equation}
This transformation places all criteria on a common scale while preserving
their preference order. Let
\(
\mathbf{v}_{i,B}
=
[v_{i,B}^{(1)},\ldots,v_{i,B}^{(M)}]^{\top}
\)
denote the normalized valuation vector of source example \(i\), and let
\(
\mathbf{V}_{B}\in[0,1]^{N\times M}
\)
be the aligned matrix obtained by stacking these vectors.

For each target domain and source budget, TAVO optimizes a nonnegative fusion
vector on the probability simplex,
\begin{equation}
\mathbf{w}
\in
\Delta^{M-1}
=
\left\{
\mathbf{w}\in\mathbb{R}^{M}:
w_m\geq0,\;
\sum_{m=1}^{M}w_m=1
\right\}.
\label{eq:weight_simplex}
\end{equation}
The fused score for source example \(i\) is
\begin{equation}
s_{i,B}(\mathbf{w})
=
\mathbf{w}^{\top}\mathbf{v}_{i,B}
=
\sum_{m=1}^{M}w_m v_{i,B}^{(m)}.
\label{eq:fused_score}
\end{equation}

For tumor segmentation, TAVO selects a global top-\(B\) subset of source cases:
\begin{equation}
\mathcal{A}_{B}(\mathbf{w})
=
\operatorname{Top}_{B}
\left(
\{(u_i,s_{i,B}(\mathbf{w}))\}_{i=1}^{N}
\right).
\label{eq:seg_subset}
\end{equation}
Simplex vertices correspond to individual criteria, while interior points fuse multiple criteria. Thus, TAVO learns a fused scoring rule rather than imposing predefined quotas over criterion families or source domains.

\subsection{Validation-Guided Multi-Fidelity Search}
\label{sec:weight_optimization}

The fused scores in Eq.~\eqref{eq:fused_score}, together with the top-\(B\) operator in Eq.~\eqref{eq:seg_subset}, define a candidate source subset for any fusion vector \(\mathbf{w}\). However, the usefulness of this subset is determined by downstream target performance, not by the criterion matrix alone. TAVO therefore scores fusion vectors through proxy training on their induced subsets and evaluation on \(\mathcal{T}_{\mathrm{val}}^\ell\). Since full training for every candidate is expensive, we use a fixed multi-fidelity protocol: (i) a coarse stage
evaluates many candidate fusion weights with short proxy-training runs, and (ii) a
refinement stage re-evaluates a small set of promising candidates with longer
proxy-training runs.

For fusion vector \(\mathbf{w}\), training seed \(\xi\), and evaluated
checkpoint \(k\), let
\begin{equation}
\theta_k(\mathbf{w},\xi)
=
\operatorname{Train}_{k}
\left(
\theta_{\mathrm{init}};
\mathcal{A}_{B}(\mathbf{w})
\cup
\mathcal{T}_{\mathrm{tr}}^\ell,
\xi
\right).
\label{eq:proxy_training}
\end{equation}
For a proxy-training length \(J\), candidate fitness is
\begin{equation}
\widehat{F}_{J}
\left(
\mathbf{w};\Xi
\right)
=
\operatorname{median}_{\xi\in\Xi}
\left[
\max_{k\in\mathcal{K}_{J}}
Q
\left(
f_{\theta_k(\mathbf{w},\xi)};
\mathcal{T}_{\mathrm{val}}^\ell
\right)
\right],
\label{eq:candidate_fitness}
\end{equation}
where \(\mathcal{K}_{J}\) is the fixed set of checkpoints evaluated within the
\(J\)-step proxy-training run, \(\Xi\) is a predetermined set of training seeds,
and \(Q\) is the target-validation segmentation metric. We use
\(J_{\mathrm{c}}\) for coarse exploration and a longer horizon
\(J_{\mathrm{r}}>J_{\mathrm{c}}\) to refine the top \(R\) coarse-stage
configurations from scratch.

The objective in Eq.~\eqref{eq:candidate_fitness} is non-smooth because of hard
top-\(B\) source selection and noisy because of stochastic network training.
We therefore use CMA-ES~\citep{hansen2003cma}, a derivative-free
population-based optimizer, to propose fusion weights. Each raw CMA-ES sample
\(\mathbf{z}\in\mathbb{R}^{M}\) is mapped to the simplex by nonnegative
normalization, \(\mathbf{w}=[\mathbf{z}]_{+}/\mathbf{1}^{\top}[\mathbf{z}]_{+}\),
with the uniform vector used when the denominator is zero. We also explicitly
evaluate all \(M\) simplex corners and the uniform fusion vector, so every individual criterion and the equal-weight mixture are included in the search. \Cref{alg:tavo} summarizes the coarse-to-refinement search for one target--budget setting.

\begin{algorithm}[t]
\caption{TAVO search for one target--budget setting}
\label{alg:tavo}
\small
\begin{algorithmic}[1]
\Require Aligned criterion matrix \(\mathbf{V}_{B}\), budget \(B\), subset
operator \(\mathcal{A}_{B}\)
\Require Proxy-training horizons \(J_{\mathrm{c}}<J_{\mathrm{r}}\), seed sets \(\Xi_{\mathrm{c}},\Xi_{\mathrm{r}}\)
\Require CMA-ES parameters \(G,\lambda,\mu,\sigma_{0}\), refinement size \(R\)
\Ensure Fusion weights \(\mathbf{w}_{B}^{\star}\) and selected subset
\(\mathcal{A}_{B}^{\star}\)

\State \(\mathcal{H}\gets\emptyset\)
\State \(\mathcal{W}_{0}\gets
\{\mathbf{e}_{m}\}_{m=1}^{M}\cup\{M^{-1}\mathbf{1}\}\)

\ForAll{\(\mathbf{w}\in\mathcal{W}_{0}\)}
    \State \(q\gets
    \widehat{F}_{J_{\mathrm{c}}}
    (\mathbf{w};\Xi_{\mathrm{c}})\)
    \State Add \((\mathbf{w},q)\) to \(\mathcal{H}\)
\EndFor

\State \(\Omega\gets
\Call{InitCMA}{M^{-1}\mathbf{1},\sigma_{0}}\)

\For{\(g=1,\ldots,G\)}
    \State
    \(\{(\mathbf{z}_{j},\mathbf{w}_{j})\}_{j=1}^{\lambda}
    \gets\Call{SampleAndMapCMA}{\Omega,\lambda}\)

    \For{\(j=1,\ldots,\lambda\)}
        \State \(q_j\gets
        \widehat{F}_{J_{\mathrm{c}}}
        (\mathbf{w}_{j};\Xi_{\mathrm{c}})\)
        \State Add \((\mathbf{w}_{j},q_j)\) to \(\mathcal{H}\)
    \EndFor

    \State \(\Omega\gets
    \Call{UpdateCMA}
    {\Omega,\{(\mathbf{z}_{j},q_j)\}_{j=1}^{\lambda},\mu}\)
\EndFor

\State \(\mathcal{C}_{R}\gets
\Call{TopDistinctWeights}{\mathcal{H},R}\)

\ForAll{\(\mathbf{w}\in\mathcal{C}_{R}\)}
    \State Refine:
    \(\widetilde q(\mathbf{w})\gets
    \widehat{F}_{J_{\mathrm{r}}}
    (\mathbf{w};\Xi_{\mathrm{r}})\)
\EndFor

\State \(\mathbf{w}_{B}^{\star}\gets
\arg\max_{\mathbf{w}\in\mathcal{C}_{R}}
\widetilde q(\mathbf{w})\)
\State \(\mathcal{A}_{B}^{\star}\gets
\mathcal{A}_{B}(\mathbf{w}_{B}^{\star})\)
\State \Return \(\mathbf{w}_{B}^{\star},\mathcal{A}_{B}^{\star}\)

\end{algorithmic}
\end{algorithm}

Here, \(\Call{TopDistinctWeights}{\mathcal{H},R}\)
returns the \(R\) distinct fusion vectors with the highest coarse-fidelity
scores. Every refinement evaluation is reinitialized from
\(\theta_{\mathrm{init}}\), rather than continued from its coarse-training
trajectory. Exact training horizons, checkpoint schedules, seed sets, CMA-ES
parameters, and proxy-training horizons are provided in Supplementary
Sec.~A.3. All search settings are fixed before test evaluation.

\subsection{Final Training and Computational Interface}
\label{sec:final_training}

After search, TAVO freezes the selected source-case manifest
\(\mathcal{A}_{B}^{\star}\), reinitializes the task model from
\(\theta_{\mathrm{init}}\), and trains on
\(\mathcal{A}_{B}^{\star}\cup\mathcal{T}_{\mathrm{tr}}^\ell\) using the same
full-training protocol as the comparison methods. During final training,
\(\mathcal{T}_{\mathrm{val}}^\ell\) is used only for checkpoint selection, and
\(\mathcal{T}_{\mathrm{te}}\) is evaluated only after the source subset,
fusion weights, and checkpoint have all been fixed.

TAVO changes neither the downstream architecture nor the task loss, and
therefore introduces no inference-time overhead. Given cached criterion
rankings, producing a fused source ordering requires
\(\mathcal{O}(NM+N\log N)\) operations in our implementation. The measured
costs of criterion warm-up, evidence extraction, criterion scoring,
multi-fidelity search, and final training are reported in
Sec.~\ref{subsec:computational_cost} and Supplementary Sec.~A.4.

\section{Experiments}
\label{sec:experimental_protocol}

\subsection{Experimental Setup}
\label{subsec:experimental_setting}

\paragraph{Benchmarks, protocols, and metrics.} We evaluate TAVO primarily on two cross-center medical image segmentation
benchmarks, with a compact natural-image classification control experiment
reported in Sec.~\ref{subsec:cross_task}. BraTS 2021 provides 1,251 annotated multimodal brain MRI training cases, with annotations for enhancing tumor (ET), tumor core (TC), and whole tumor (WT) \citep{baid2021brats, menze2015brats,bakas2017advancing}. MAMA-MIA contains 1,506 pre-treatment breast DCE-MRI cases from four centers: NACT, I-SPY1, I-SPY2, and Duke \citep{garrucho2025mamamia}. OfficeHome contains 15,588 images from 65 categories across Art, Clipart, Product, and Real World \citep{venkateswara2017deep}. We use dataset-provided center labels and official case-mapping files to define domains, and use fixed split manifests for all reported experiments.

For segmentation, we use leave-one-domain-out evaluation: one center/domain is held out as the target, and all remaining domains form the labeled source pool. Target cases are split into disjoint training, validation, and test sets, \(\mathcal{T}_{\mathrm{tr}}\), \(\mathcal{T}_{\mathrm{val}}\), and \(\mathcal{T}_{\mathrm{te}}\). The training split supervises candidate and final training and provides target evidence for target-conditioned criteria; the validation split selects TAVO fusion weights and model checkpoints; and the test split is accessed only for final reporting.

For BraTS, we report 3D case-level Macro Dice averaged over ET, TC, and WT. For MAMA-MIA, we report foreground Dice. For both benchmarks, we evaluate source budgets \(B\in\{50,150,250,500\}\) cases. Each setting is repeated with three final-training seeds, and dataset-level summaries are unweighted averages over the four held-out target domains. Exact target split sizes, source-pool sizes, and source-budget percentages are provided in Supplementary Sec.~B.1.

\paragraph{Baselines and fairness.}
We compare TAVO with three groups of methods:
(i) non-budgeted references, including Target-only, Full Source-only,
and Target + Full Source;
(ii) budget-matched source-curation baselines, including Random and the eight
fixed single-criterion selectors spanning the target-conditioned and
source-structural criteria described in Sec.~\ref{sec:rank_fusion}; and
(iii) model-centric adaptation baselines for segmentation, including DANN,
MMD, ADVENT, and SE-ASA \citep{ganin2016dann,long2015dan,vu2019advent,feng2023seasa}.
All budgeted selectors and budget-matched adaptation baselines use exactly \(B\) source examples together with the same labeled target training support. Budget-matched adaptation baselines are trained on a random \(B\)-case source subset, so adaptation is evaluated under the same source-cardinality constraint as the selection methods. Full-source references are reported only as non-budgeted references.

\paragraph{Implementation and reporting.} Within each benchmark, all methods share the same preprocessing, backbone, optimizer schedule, training length, and checkpoint-selection rule. BraTS and MAMA-MIA use nnU-Net preprocessing, 2D slice-level training, and reconstructed 3D case-level evaluation \citep{isensee2021nnunet}. BraTS uses EfficientViT-Seg-L1 initialized from ADE20K-pretrained weights \citep{cai2023efficientvit,zhou2017scene}, which provides an efficient pretrained segmentation backbone for repeated subset training and valuation search; MAMA-MIA uses the nnUNet v2 2D configuration with three DCE channels, allowing us to evaluate TAVO on a standardized medical segmentation pipeline \citep{isensee2021nnunet}. Final training uses 20 epochs for BraTS and 100 epochs for MAMA-MIA.

For each held-out target, we train one early Target + Full Source warm-up model and use its frozen checkpoint only to cache representations and gradients for source valuation. Candidate and final models are reinitialized from the same task-specific initialization and trained from scratch. Search hyperparameters and proxy-training lengths are fixed before test evaluation and reported in Supplementary Sec.~A.3. The source budget constrains candidate and final training; the one-time full-pool warm-up, evidence extraction, and criterion scoring costs are reported in Sec.~\ref{subsec:computational_cost}. Additional implementation and metric details are provided in Supplementary Secs.~B.2--B.3.

\subsection{Main Results under Matched Source Budgets}
\label{subsec:main_results}

\begin{figure*}[t]
  \centering
  \includegraphics[width=\textwidth]{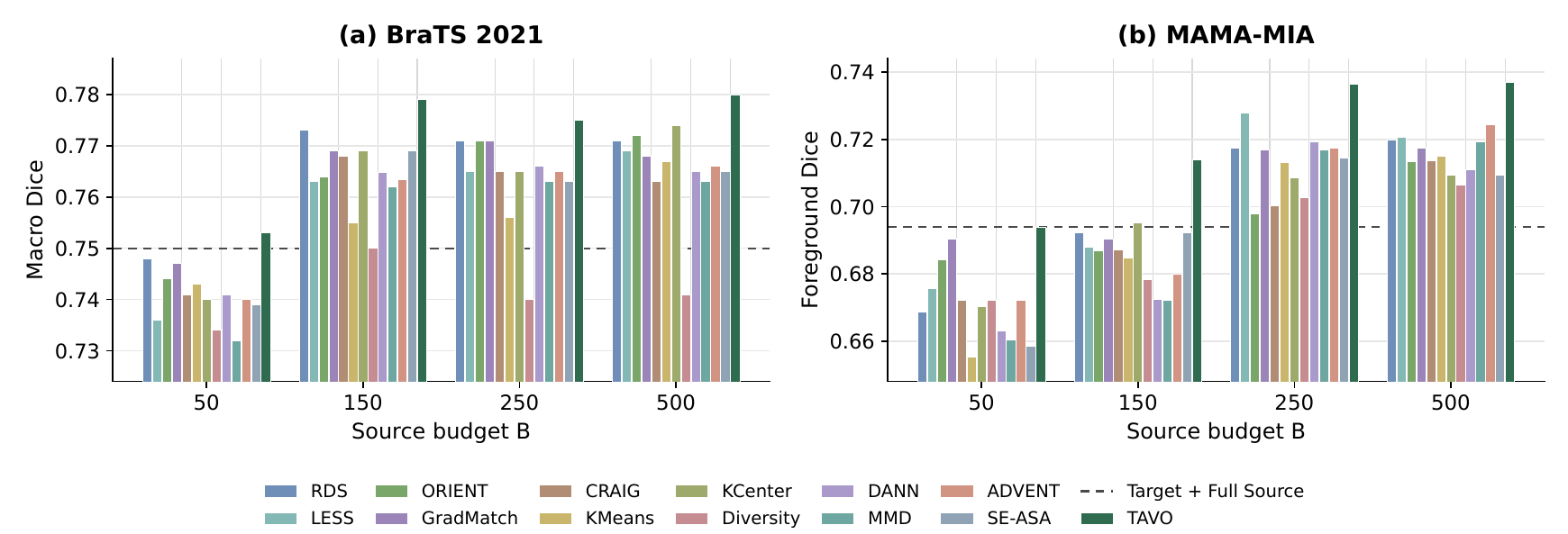}
  \caption{
  Target-averaged performance across source budgets on BraTS 2021 and MAMA-MIA. Each method is evaluated under matched source budgets; the dashed line denotes
the non-budgeted Target + Full Source reference.}
  \label{fig:budget_bars}
\end{figure*}

\paragraph{Overall performance across budgets.}
\Cref{fig:budget_bars} summarizes target-averaged segmentation performance under matched source budgets. Across BraTS 2021 and MAMA-MIA, TAVO remains among the strongest methods as the source budget varies, despite using a single adaptive fusion strategy rather than a fixed target-conditioned, source-structural, or adaptation objective. On both medical benchmarks, TAVO also outperforms the Target + Full Source reference in most budget settings while using substantially fewer external cases during final training. This shows that cross-center source data are not uniformly beneficial:
under domain shift, indiscriminately adding all available source cases
can be suboptimal, and the composition of the source set can matter as
much as its size. Complete per-target results across all source budgets are provided in Supplementary Sec.~C.1.

\newcommand{\valstd}[2]{#1\raisebox{-0.04ex}{\fontsize{4.4pt}{4.6pt}\selectfont$\pm#2$}}
\newcommand{\bvalstd}[2]{\textbf{#1}\raisebox{-0.04ex}{\fontsize{4.4pt}{4.6pt}\selectfont$\pm#2$}}
\newcommand{\cat}[2]{\multirow{#1}{*}{\scriptsize\shortstack[l]{#2}}}

\newcommand{\uvalstd}[2]{%
  \begingroup
  \setbox0=\hbox{\valstd{#1}{#2}}%
  \leavevmode
  \ooalign{%
    \copy0\cr
    \hidewidth\raisebox{-0.5ex}{\rule{\wd0}{0.28pt}}\hidewidth\cr
  }%
  \endgroup
}
\newcommand{\bnum}[1]{\textbf{#1}}
\newcommand{\unum}[1]{%
  \begingroup
  \setbox0=\hbox{#1}%
  \leavevmode
  \ooalign{%
    \copy0\cr
    \hidewidth\raisebox{-0.5ex}{\rule{\wd0}{0.28pt}}\hidewidth\cr
  }%
  \endgroup
}

\newcommand{\gsep}{\rule[-0.25ex]{0.30pt}{1.55ex}}

\begin{table*}[t]
\centering
\caption{
Per-target segmentation results at the source budget ($B=150$).
The top block contains non-budgeted references; LGG and GBM
denote TCGA-LGG and TCGA-GBM; Boldface and underlining indicate
the best and second-best budget-matched results, respectively.
}
\label{tab:seg_representative_targets}
\fontsize{7.0pt}{7.7pt}\selectfont
\setlength{\tabcolsep}{1.1pt}
\renewcommand{\arraystretch}{0.96}
\begin{tabularx}{\textwidth}{@{}l@{\hspace{3pt}\gsep\hspace{3pt}}l*{5}{Y}@{\hspace{3pt}}c@{\hspace{3pt}}*{5}{Y}@{}}
\toprule
\multirow{2}{*}{Category}
& \multirow{2}{*}{Method}
& \multicolumn{5}{c}{BraTS 2021 Macro Dice $\uparrow$}
& \gsep
& \multicolumn{5}{c}{MAMA-MIA Foreground Dice $\uparrow$} \\
\cmidrule(lr){3-7}\cmidrule(lr){9-13}
&
& C4 & C5 & LGG & GBM & Avg.
& \gsep
& NACT & I-SPY1 & Duke & I-SPY2 & Avg. \\
\midrule
\cat{3}{Non-budgeted\\refs.}
& Target-only
& \valstd{70.7}{0.3}
& \valstd{77.7}{0.1}
& \valstd{56.5}{0.3}
& \valstd{82.4}{0.2}
& 71.8
& \gsep
& \valstd{53.6}{0.7}
& \valstd{66.1}{0.5}
& \valstd{56.3}{0.1}
& \valstd{74.7}{0.2}
& 62.7 \\

& Full Source-only 
& \valstd{76.0}{0.6}
& \valstd{73.4}{0.2}
& \valstd{59.5}{0.7}
& \valstd{79.0}{0.4}
& 72.0
& \gsep
& \valstd{71.5}{0.4}
& \valstd{69.6}{0.3}
& \valstd{56.0}{0.4}
& \valstd{62.5}{0.8}
& 64.9 \\

& Target + Full Source
& \valstd{76.1}{0.5}
& \valstd{79.1}{0.3}
& \valstd{61.6}{0.4}
& \valstd{83.2}{0.2}
& 75.0
& \gsep
& \valstd{72.1}{0.5}
& \valstd{70.6}{0.2}
& \valstd{59.4}{0.5}
& \valstd{75.4}{0.6}
& 69.4 \\

\midrule
\cat{1}{Random}
& Random
& \valstd{75.9}{1.1}
& \valstd{82.4}{0.8}
& \valstd{60.9}{0.6}
& \valstd{83.4}{0.3}
& 75.6
& \gsep
& \valstd{67.8}{0.5}
& \valstd{69.2}{0.6}
& \valstd{63.5}{0.3}
& \valstd{76.3}{0.8}
& 69.2 \\
\midrule
\cat{4}{Target-\\conditioned}
& RDS~\citep{ruder2017select,ma2019domain}
& \valstd{76.5}{0.1}
& \valstd{81.4}{0.2}
& \bvalstd{63.8}{1.8}
& \uvalstd{84.6}{0.2}
& 76.6
& \gsep
& \valstd{70.3}{0.8}
& \valstd{65.8}{0.2}
& \bvalstd{64.4}{0.4}
& \valstd{76.4}{0.4}
& 69.2 \\

& LESS~\citep{xia2024less}
& \valstd{77.9}{1.0}
& \valstd{80.8}{0.4}
& \valstd{62.6}{0.6}
& \valstd{83.8}{0.2}
& 76.3
& \gsep
& \valstd{69.9}{0.3}
& \valstd{67.8}{0.2}
& \valstd{62.4}{0.3}
& \valstd{75.1}{0.4}
& 68.8 \\

& ORIENT~\citep{karanam2022orient}
& \valstd{76.9}{1.3}
& \uvalstd{82.7}{0.4}
& \valstd{61.7}{0.4}
& \valstd{84.4}{0.0}
& 76.4
& \gsep
& \valstd{67.4}{0.3}
& \uvalstd{71.1}{0.2}
& \valstd{60.0}{0.7}
& \valstd{76.2}{0.5}
& 68.7 \\

& GradMatch~\citep{killamsetty2021gradmatch}
& \valstd{78.5}{1.7}
& \valstd{82.3}{0.1}
& \valstd{62.4}{1.9}
& \valstd{84.5}{0.4}
& \unum{76.9}
& \gsep
& \valstd{68.6}{0.5}
& \valstd{68.9}{0.3}
& \valstd{62.5}{0.5}
& \valstd{76.2}{0.6}
& 69.1 \\

\midrule
\cat{4}{Source-\\structural}
& CRAIG~\citep{mirzasoleiman2020craig}
& \valstd{78.8}{1.3}
& \valstd{82.2}{0.4}
& \valstd{61.8}{2.0}
& \valstd{84.2}{0.0}
& 76.8
& \gsep
& \valstd{69.6}{0.2}
& \valstd{67.7}{0.4}
& \valstd{61.9}{0.4}
& \valstd{75.7}{0.2}
& 68.7 \\

& KMeans~\citep{lloyd1982least}
& \valstd{77.5}{1.3}
& \valstd{82.2}{0.3}
& \valstd{58.0}{1.4}
& \valstd{84.3}{0.1}
& 75.5
& \gsep
& \valstd{66.0}{1.0}
& \valstd{69.3}{0.4}
& \valstd{62.4}{1.0}
& \valstd{76.2}{0.5}
& 68.5 \\

& KCenter~\citep{sener2018coreset}
& \valstd{78.7}{1.1}
& \valstd{82.4}{0.1}
& \valstd{62.4}{0.6}
& \valstd{84.0}{0.2}
& 76.8
& \gsep
& \uvalstd{71.0}{0.2}
& \valstd{68.1}{0.2}
& \valstd{63.2}{0.5}
& \valstd{75.8}{1.1}
& \unum{69.5} \\

& Diversity~\citep{gonzalez1985clustering}
& \valstd{79.0}{1.7}
& \valstd{80.2}{0.1}
& \valstd{57.5}{1.1}
& \valstd{83.2}{0.2}
& 75.0
& \gsep
& \valstd{65.9}{0.9}
& \valstd{68.7}{0.3}
& \valstd{60.6}{1.3}
& \valstd{76.2}{0.3}
& 67.8 \\

\midrule
\cat{4}{Budget-\\matched DA}
& DANN~\citep{ganin2016dann}
& \valstd{78.3}{2.0}
& \valstd{82.2}{0.0}
& \valstd{61.0}{2.4}
& \valstd{84.4}{0.3}
& 76.5
& \gsep
& \valstd{67.0}{1.1}
& \valstd{66.2}{0.1}
& \valstd{61.6}{0.2}
& \valstd{74.2}{0.6}
& 67.3 \\

& MMD~\citep{long2015dan}
& \valstd{78.2}{0.7}
& \valstd{81.7}{0.9}
& \valstd{60.7}{1.5}
& \valstd{84.2}{0.4}
& 76.2
& \gsep
& \valstd{68.1}{0.9}
& \valstd{64.3}{0.3}
& \valstd{62.7}{0.6}
& \valstd{73.8}{0.4}
& 67.2 \\

& ADVENT~\citep{vu2019advent}
& \valstd{77.6}{2.2}
& \valstd{82.1}{0.0}
& \valstd{61.2}{0.4}
& \valstd{84.5}{0.1}
& 76.4
& \gsep
& \valstd{67.5}{0.4}
& \valstd{65.1}{0.2}
& \valstd{62.9}{1.0}
& \uvalstd{76.5}{0.4}
& 68.0 \\

& SE-ASA~\citep{feng2023seasa}
& \bvalstd{80.2}{0.6}
& \valstd{82.6}{0.2}
& \valstd{60.9}{0.7}
& \valstd{83.9}{0.1}
& \unum{76.9}
& \gsep
& \valstd{69.4}{0.3}
& \valstd{68.9}{0.5}
& \valstd{63.2}{0.6}
& \valstd{75.4}{0.1}
& 69.2 \\

\midrule
\cat{1}{Ours}
& \textbf{TAVO}
& \uvalstd{79.9}{0.6}
& \bvalstd{83.0}{0.5}
& \uvalstd{63.7}{0.9}
& \bvalstd{84.9}{0.1}
& \bnum{77.9}
& \gsep
& \bvalstd{72.9}{0.3}
& \bvalstd{71.2}{0.2}
& \uvalstd{64.2}{0.5}
& \bvalstd{77.3}{0.1}
& \bnum{71.4} \\

\bottomrule
\end{tabularx}
\end{table*}

\begin{figure}[t]
  \centering
  \includegraphics[width=\linewidth]{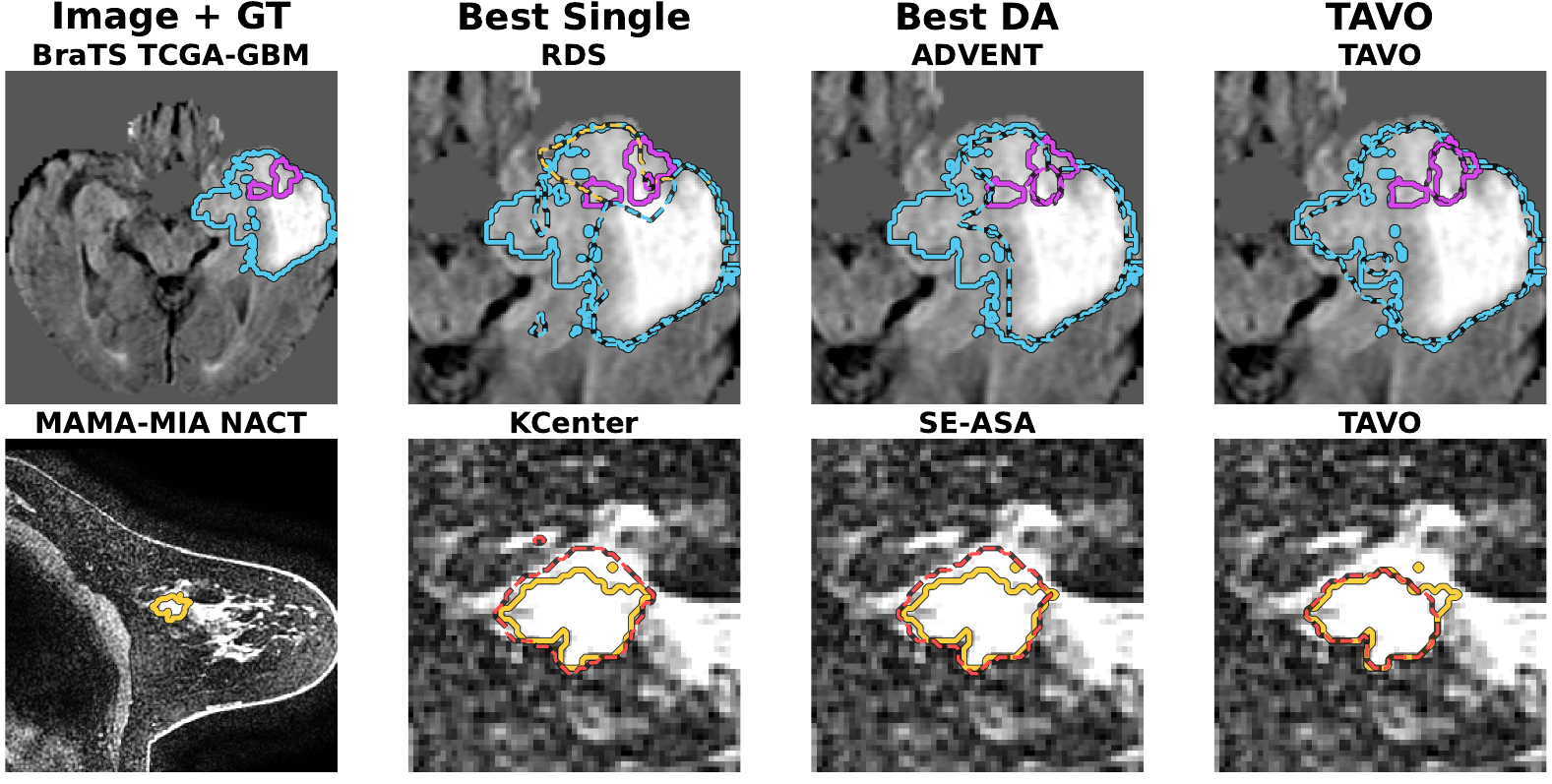}
  \caption{
  Qualitative segmentation examples at the 150-case source budget ($B=150$). Solid contours denote ground truth and dashed contours denote predictions. For BraTS, WT, TC, and ET are shown in yellow, cyan, and magenta, respectively. For MAMA-MIA, ground-truth and predicted tumor contours are shown in yellow and red, respectively.
  }
  \label{fig:qualitative_examples}
\end{figure}

\paragraph{Representative tumor segmentation results.}
\Cref{tab:seg_representative_targets} reports per-target results at the
representative 150-case source budget. TAVO achieves 77.9 Macro Dice on BraTS
and 71.4 foreground Dice on MAMA-MIA, improving over Random by 2.3/2.2 points,
the strongest fixed selector by 1.0/1.9 points, and the strongest
budget-matched adaptation baseline by 1.0/2.2 points, respectively. The gain is
not driven by a single favorable target: TAVO ranks first on five of the eight
tumor-segmentation targets and second on the remaining three, while the best
competing fixed selector changes across centers. TAVO also exceeds the non-budgeted Target + Full Source reference
on all eight targets, improving average Dice by 2.9/2.0 points on
BraTS/MAMA-MIA while selecting fewer source cases. Qualitative
examples in \Cref{fig:qualitative_examples} provide representative
contour-level comparisons. Paired target-level tests show significant gains over the strongest
fixed selector on BraTS and MAMA-MIA ($p=.029/.025$).
A controlled BraTS nnU-Net comparison further supports backbone
robustness, with TAVO achieving 82.5 Dice versus 81.5 for
Target + Full Source (Supplementary Secs.~C.4--C.5).

\subsection{Computational Cost and Practical Trade-offs}
\label{subsec:computational_cost}

\begin{table}[t]
\centering
\caption{Measured GPU-hours for representative tumor-segmentation settings at budget (\(B=150\)) on an NVIDIA A40 GPU. Costs are reported for one final-training seed.}
\label{tab:compute_cost}
\footnotesize
\setlength{\tabcolsep}{3.0pt}
\renewcommand{\arraystretch}{1.02}
\begin{tabular}{lcc}
\toprule
Cost component
& BraTS
& MAMA-MIA\\
\midrule
Criterion warm-up checkpoint
& 7.8 & 6.9 \\
Selection evidence extraction
& 5.6 & 4.1 \\
Criterion scoring
& 0.2 & 0.2 \\
Multi-fidelity search
& 18.5 & 15.9 \\
\midrule
Curation overhead
& 32.1 & 27.1 \\
Final training
& 6.0 & 5.7 \\
TAVO end-to-end
& 38.1 & 32.8 \\
Target + Full Source training
& 32.0 & 28.6 \\
\midrule
End-to-end / full-source training
& $1.19\times$ & $1.15\times$ \\
\bottomrule
\end{tabular}
\end{table}

\Cref{tab:compute_cost} reports TAVO's one-time development cost at $B=150$.
The source budget constrains candidate and final training, while warm-up and
evidence extraction access the full source pool once per target.
Total cost is $1.19\times/1.15\times$ one Target + Full Source run on
BraTS/MAMA-MIA, while remaining below a full-fidelity eight-selector sweep
(61.6/56.8 GPU-hours). This overhead is confined to development: once the
source manifest is fixed, TAVO adds no deployment- or inference-time cost.
Search-budget sensitivity is reported in Supplementary Sec.~C.6.

\subsection{Target- and Budget-Specific Fusion Behavior}
\label{subsec:fusion_behavior}

\begin{figure}[t]
  \centering
  \includegraphics[width=\linewidth]{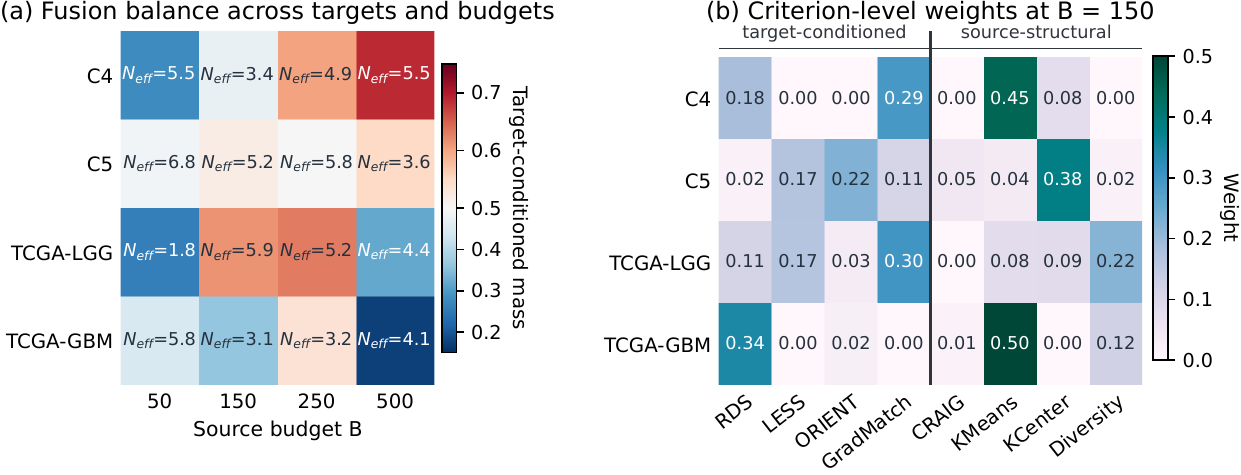}
  \caption{
  Target- and budget-specific fusion behavior on BraTS.
  }
  \label{fig:fusion_behavior}
\end{figure}

\paragraph{Learned fusion weights.}
\Cref{fig:fusion_behavior} shows that TAVO does not learn a single target-invariant weighting policy. Both the target-conditioned weight mass and the effective number of active criteria, \(N_{\mathrm{eff}}=1/\sum_{m=1}^{M}w_m^2\), vary across held-out centers and source budgets, indicating that TAVO adapts the trade-off between target relevance and source-structural coverage to each transfer setting. At the representative 150-case source budget, the criterion-level profiles also differ across targets, consistent with \Cref{tab:seg_representative_targets}, where the strongest fixed selector changes by center. We therefore interpret the learned weights as descriptive fusion profiles rather than causal estimates of criterion importance.

\subsection{Ablation Studies}
\label{subsec:ablation}

\begin{table}[t]
\centering
\caption{
Ablation at the representative source budget used for each benchmark. Values are averaged over
the four target domains.
}
\label{tab:ablation}
\footnotesize
\setlength{\tabcolsep}{3.0pt}
\renewcommand{\arraystretch}{1.5}
\begin{tabular}{lcc}
\toprule
Variant
& BraTS
& MAMA-MIA\\
\midrule
Source-structural only
& 76.2 & 69.2 \\

Target-conditioned only
& \underline{76.3} & \underline{70.3} \\

Uniform fusion
& 76.1 & 69.4 \\

Val.-selected single
& 75.9 & 68.5 \\

Dirichlet random search
& 75.2 & 67.9 \\

Shuffled-ranking CMA-ES
& 76.0 & 69.5 \\

\midrule
Full TAVO
& \textbf{77.9}
& \textbf{71.4} \\
\bottomrule
\end{tabular}
\end{table}

\Cref{tab:ablation} evaluates the contributions of complementary
valuation signals, adaptive fusion, and guided search.
Full TAVO outperforms the strongest single-family variant by
1.6/1.1 points and uniform fusion by 1.8/2.0 points on
BraTS/MAMA-MIA, respectively, indicating that both criterion
complementarity and adaptive weighting are beneficial.

Compared with Val.-selected single and Dirichlet random search,
TAVO improves performance by 2.0/2.9 and 2.7/3.5 points,
respectively. Moreover, shuffling criterion rankings while
retaining CMA-ES reduces performance to 76.0/69.5,
suggesting that optimization alone cannot compensate for
uninformative valuation signals.
Per-target results are in Supplementary Sec.~C.3.

\subsection{Cross-Task Control on OfficeHome}
\label{subsec:cross_task}

\begin{table}[t]
\centering
\caption{
OfficeHome accuracy at \(b=15\) selected source images per class, corresponding to \(B=975\) total selected source images. The top block contains non-budgeted references; bold and
underline indicate the best and second-best budget-matched results,
respectively.
}
\label{tab:officehome_b15_acc}
\fontsize{6.4pt}{7.0pt}\selectfont
\setlength{\tabcolsep}{1.4pt}
\renewcommand{\arraystretch}{0.95}
\resizebox{\linewidth}{!}{%
\begin{tabular}{lccccc}
\toprule
Method & Art & Clipart & Product & Real & Avg. \\
\midrule
Target-only
& \valstd{43.9}{1.5}
& \valstd{26.3}{0.4}
& \valstd{56.6}{1.8}
& \valstd{51.2}{1.6}
& 44.5 \\

Source-only Full
& \valstd{60.3}{1.1}
& \valstd{40.3}{0.8}
& \valstd{75.2}{0.2}
& \valstd{75.5}{0.7}
& 62.8 \\

Target + Full Source
& \valstd{63.3}{1.0}
& \valstd{49.8}{0.4}
& \valstd{78.7}{0.5}
& \valstd{77.0}{0.5}
& 67.2 \\

\midrule

Random
& \valstd{56.6}{1.3}
& \valstd{44.0}{0.9}
& \valstd{71.1}{0.3}
& \valstd{68.3}{1.0}
& 60.0 \\

\midrule

KMeans~\citep{lloyd1982least}
& \valstd{57.7}{0.3}
& \valstd{44.7}{1.1}
& \valstd{72.3}{1.5}
& \valstd{68.0}{0.5}
& 60.7 \\

KCenter~\citep{sener2018coreset}
& \valstd{49.5}{0.7}
& \valstd{40.2}{0.2}
& \valstd{67.7}{1.0}
& \valstd{61.6}{1.6}
& 54.7 \\

Facility Location~\citep{wei2015submodularity}
& \valstd{58.0}{1.2}
& \valstd{43.9}{0.8}
& \valstd{72.4}{0.8}
& \valstd{68.9}{0.5}
& 60.8 \\

GradMatch~\citep{killamsetty2021gradmatch}
& \valstd{58.6}{1.2}
& \valstd{44.3}{0.9}
& \valstd{73.6}{0.6}
& \valstd{66.4}{1.3}
& 60.7 \\

\midrule

CRAIG~\citep{mirzasoleiman2020craig}
& \valstd{55.4}{0.5}
& \valstd{44.6}{0.3}
& \valstd{72.7}{0.7}
& \uvalstd{69.2}{0.4}
& 60.5 \\

GLISTER~\citep{killamsetty2021glister}
& \uvalstd{59.7}{1.0}
& \valstd{43.9}{1.3}
& \uvalstd{74.9}{0.7}
& \valstd{68.1}{0.7}
& 61.7 \\

ORIENT~\citep{karanam2022orient}
& \bvalstd{60.6}{0.4}
& \uvalstd{46.1}{1.2}
& \valstd{73.5}{0.4}
& \valstd{68.4}{0.4}
& \unum{62.2} \\

MomentMatching~\citep{gretton2012kernel,long2015dan}
& \valstd{58.2}{1.1}
& \valstd{44.1}{0.6}
& \valstd{73.4}{0.1}
& \valstd{66.8}{0.9}
& 60.6 \\

\midrule

DANN~\citep{ganin2016dann}
& \valstd{57.9}{0.2}
& \valstd{41.6}{0.3}
& \valstd{70.9}{0.9}
& \valstd{68.9}{0.8}
& 59.8 \\

MMD~\citep{long2015dan}
& \valstd{58.5}{0.6}
& \valstd{41.2}{0.7}
& \valstd{70.6}{0.5}
& \valstd{68.7}{0.4}
& 59.8 \\

CORAL~\citep{sun2016coral}
& \valstd{57.1}{0.3}
& \valstd{41.6}{0.6}
& \valstd{70.7}{0.2}
& \valstd{68.9}{0.6}
& 59.6 \\

CDAN~\citep{long2018cdan}
& \valstd{58.3}{0.5}
& \valstd{42.0}{0.5}
& \valstd{70.9}{1.1}
& \valstd{69.1}{0.4}
& 60.1 \\

\midrule

\textbf{TAVO}
& \uvalstd{59.7}{1.0}
& \bvalstd{46.5}{0.3}
& \bvalstd{75.2}{0.7}
& \bvalstd{69.5}{0.7}
& \bnum{62.7} \\

\bottomrule
\end{tabular}
}
\end{table}

To test whether the same source-curation interface transfers beyond medical segmentation, we include a compact OfficeHome classification control experiment \citep{venkateswara2017deep}. Each task pools three domains as the source pool and holds out the remaining domain as the target, with three target-training and two validation images per class; source budgets select \(b\) source images per class. Since OfficeHome has 65 categories, the total source budget is \(B=65b\). All methods use an ImageNet-pretrained ResNet-50 and are trained for 30 epochs.  TAVO uses the same rank-fusion interface as in segmentation, but replaces the segmentation criterion library with classification counterparts: MomentMatching~\citep{gretton2012kernel,long2015dan}, GradMatch~\citep{killamsetty2021gradmatch}, GLISTER~\citep{killamsetty2021glister}, and ORIENT~\citep{karanam2022orient} as target-conditioned criteria, and CRAIG~\citep{mirzasoleiman2020craig}, KMeans~\citep{lloyd1982least}, KCenter~\citep{sener2018coreset}, and Facility Location~\citep{wei2015submodularity} as source-structural criteria. We compare against Random, fixed selectors, and budget-matched classification adaptation baselines DANN~\citep{ganin2016dann}, MMD~\citep{long2015dan}, CORAL~\citep{sun2016coral}, and CDAN~\citep{long2018cdan}. At \(b=15\) in \Cref{tab:officehome_b15_acc}, TAVO achieves the strongest budget-matched average accuracy
(62.7\%), outperforming Random, the strongest fixed selector, and the
strongest budget-matched adaptation baseline by 2.7/0.5/2.6 points,
respectively. These results provide supporting evidence that TAVO is not tied to a particular segmentation architecture, medical imaging modality, or evaluation metric.

\section{Conclusion}
\label{sec:conclusion}

We introduced TAVO, a model-external source-curation workflow for budgeted cross-center tumor segmentation. TAVO rank-normalizes complementary valuation signals and learns target- and budget-specific fusion weights from held-out target validation, producing a source-case manifest for an otherwise unchanged training and inference pipeline.

Across BraTS 2021 and MAMA-MIA, TAVO consistently improves over random selection and strong fixed selectors under matched source budgets, and in medical settings exceeds Target + Full Source while using fewer external cases during final training. OfficeHome further supports transfer beyond MRI segmentation. These results highlight source composition as a first-class model-development decision under domain shift. TAVO requires labeled target validation, a full-pool warm-up, and target–budget-specific search; future work should reduce these costs, study robustness to smaller or shifted validation support, and validate selected-case composition prospectively.

\clearpage
\section*{Acknowledgments}
This work was supported by the National Science Foundation (NSF) under grants 2613950 and 2613761.

{
    \small
    \bibliographystyle{ieeenat_fullname}
    \bibliography{main}
}

\clearpage
\begingroup

\setcounter{table}{0}
\renewcommand{\thetable}{S\arabic{table}}
\renewcommand{\theHtable}{supp.\arabic{table}}
\setcounter{figure}{0}
\renewcommand{\thefigure}{S\arabic{figure}}
\renewcommand{\theHfigure}{supp.\arabic{figure}}
\setcounter{equation}{0}
\renewcommand{\theequation}{S\arabic{equation}}
\renewcommand{\theHequation}{supp.\arabic{equation}}
\let\valstd\relax
\let\bvalstd\relax
\let\uvalstd\relax
\let\bnum\relax
\let\unum\relax
\setlength{\textfloatsep}{6pt plus 1pt minus 2pt}
\setlength{\dbltextfloatsep}{6pt plus 1pt minus 2pt}
\setlength{\floatsep}{6pt plus 1pt minus 2pt}
\setlength{\dblfloatsep}{6pt plus 1pt minus 2pt}
\setlength{\intextsep}{6pt plus 1pt minus 2pt}

\makeatletter
\setlength{\@fptop}{0pt}
\setlength{\@fpsep}{8pt plus 2pt minus 2pt}
\setlength{\@fpbot}{0pt plus 1fil}
\setlength{\@dblfptop}{0pt}
\setlength{\@dblfpsep}{8pt plus 2pt minus 2pt}
\setlength{\@dblfpbot}{0pt plus 1fil}
\makeatother
\renewcommand{\floatpagefraction}{0.25}
\renewcommand{\dblfloatpagefraction}{0.25}

\renewcommand{\topfraction}{0.95}
\renewcommand{\dbltopfraction}{0.95}
\renewcommand{\textfraction}{0.05}
\setcounter{topnumber}{5}
\setcounter{dbltopnumber}{5}
\newcommand{\valstd}[2]{#1\raisebox{-0.04ex}{\fontsize{5.0pt}{5.2pt}\selectfont$\pm#2$}}
\newcommand{\bvalstd}[2]{\textbf{#1}\raisebox{-0.04ex}{\fontsize{4.4pt}{4.6pt}\selectfont$\pm#2$}}
\newcommand{\uvalstd}[2]{%
  \begingroup
  \setbox0=\hbox{\valstd{#1}{#2}}%
  \leavevmode
  \ooalign{%
    \copy0\cr
    \hidewidth\raisebox{-0.5ex}{\rule{\wd0}{0.28pt}}\hidewidth\cr
  }%
  \endgroup
}
\newcommand{\bnum}[1]{\textbf{#1}}
\newcommand{\unum}[1]{%
  \begingroup
  \setbox0=\hbox{#1}%
  \leavevmode
  \ooalign{%
    \copy0\cr
    \hidewidth\raisebox{-0.5ex}{\rule{\wd0}{0.28pt}}\hidewidth\cr
  }%
  \endgroup
}

\maketitlesupplementary
\appendix
\section{Criterion and Search Details}
\label{supp:criterion_search_details}

This section provides criterion-adapter, warm-up, search,
and cost details referenced by Secs.~3.2--3.4 of the main
paper. Unless otherwise stated, the source-selection unit for
tumor segmentation is a patient case. For TAVO, the target validation
split is used only for fusion and checkpoint selection, and
final models are trained from scratch on the selected source
manifest together with the labeled target training support.

\subsection{Criterion Adapters and Cached Evidence}
\label{supp:criterion_adapters}

TAVO combines heterogeneous criteria through a common source-ranking interface.
Each criterion returns either a scalar utility for every source example or an
ordered prefix generated by a greedy selection procedure. Score-based criteria
are sorted by descending utility, with tied utilities assigned their average
rank. Ordering-based criteria retain their native greedy order. Examples outside
a returned prefix receive a common tail rank below every listed source example.
Criteria whose native output depends on the requested source budget are
recomputed separately for each budget \(B\).

\paragraph{Selection units.}
For tumor segmentation, the source-selection unit is a complete patient case.
When a criterion produces slice-level features, scores, or gradients, these
values are aggregated to the case level before ranking. Selecting a case includes
all associated slices used by the downstream learner. For the OfficeHome control,
the selection unit is one image and the final source manifest is class-balanced.

\paragraph{Criterion order for fusion weights.}
All score files and fusion-weight vectors use the criterion order in
\Cref{tab:supp_score_file_order}. A one-hot vector in this order recovers the
corresponding fixed selector. The main paper groups criteria by semantic family;
the ordering below is used only for reproducible score-file alignment and
manifest auditing.

\begin{table}[ht]
\centering
\caption{
Criterion order used for score-file alignment and fusion-weight vectors.
}
\label{tab:supp_score_file_order}
\small
\setlength{\tabcolsep}{4pt}
\renewcommand{\arraystretch}{1.08}
\begin{adjustbox}{max width=\linewidth}
\begin{tabular}{clll}
\toprule
Position & Key & Criterion & Family in the paper \\
\midrule
1 & \texttt{rds} & RDS & target-conditioned \\
2 & \texttt{less} & LESS & target-conditioned \\
3 & \texttt{orient} & ORIENT & target-conditioned \\
4 & \texttt{gradmatch} & GradMatch & target-conditioned \\
5 & \texttt{craig} & CRAIG & source-structural \\
6 & \texttt{kmeans} & KMeans & source-structural \\
7 & \texttt{kcenter} & KCenter & source-structural \\
8 & \texttt{diversity} & Diversity & source-structural \\
\bottomrule
\end{tabular}
\end{adjustbox}
\end{table}

\paragraph{Selection-time evidence.}
All applicable criteria for a held-out target share the same frozen warm-up
checkpoint and cached evidence. For BraTS, feature evidence is extracted from
the EfficientViT-Seg-L1 backbone feature map and global-average pooled at the
slice level. Gradient evidence is computed with respect to the final
segmentation-head parameters. Slice-level evidence is then aggregated to the
patient-case level before ranking source cases. For MAMA-MIA, feature evidence
consists of cached nnU-Net case embeddings from the pool/query embedding files,
and gradient evidence uses CountSketch-projected gradients from the nnU-Net
segmentation head. For OfficeHome, feature evidence is extracted from the
ImageNet-pretrained ResNet-50 representation used by the classification control.

\paragraph{Segmentation criteria.}
The segmentation library contains four target-conditioned criteria and four
source-structural criteria.

\begin{itemize}
    \item \textbf{RDS} ranks source cases by representation similarity between
    each case and the labeled target-support representation
    \citep{ruder2017select,ma2019domain}.

    \item \textbf{LESS} ranks source cases by target-conditioned gradient
    similarity, estimating whether a source case induces an update direction
    compatible with the labeled target objective \citep{xia2024less}.

    \item \textbf{ORIENT} uses target-query submodular mutual information to
    balance source--target relevance and source-set coverage
    \citep{karanam2022orient}. In our adapter, the query set is the labeled
    target training support and the ground set is the candidate source pool.

    \item \textbf{GradMatch} greedily constructs a source subset whose aggregate
    gradient approximates a target-conditioned reference gradient
    \citep{killamsetty2021gradmatch}.

    \item \textbf{CRAIG} constructs a source-gradient coreset that approximates
    full-source optimization behavior \citep{mirzasoleiman2020craig}.

    \item \textbf{KMeans} clusters source cases in cached representation space
    and selects cases nearest to cluster centroids \citep{lloyd1982least}.

    \item \textbf{KCenter} applies farthest-first traversal in representation
    space to improve source coverage \citep{sener2018coreset}.

    \item \textbf{Diversity} applies greedy max--min diversification in
    representation space to favor mutually dispersed cases
    \citep{gonzalez1985clustering}.
\end{itemize}

\paragraph{Classification control criteria.}
The OfficeHome control uses the same rank-fusion interface, but ranks source
images rather than patient cases. For each class, criteria rank images from the
pooled source domains, and TAVO selects the top \(b\) images per class, yielding
\(B=Cb\) selected source images for \(C=65\) classes. The target-conditioned
criteria are MomentMatching \citep{gretton2012kernel,long2015dan},
GradMatch \citep{killamsetty2021gradmatch}, GLISTER
\citep{killamsetty2021glister}, and ORIENT \citep{karanam2022orient}; the
source-structural criteria are CRAIG \citep{mirzasoleiman2020craig},
KMeans \citep{lloyd1982least}, KCenter \citep{sener2018coreset}, and
Facility Location \citep{wei2015submodularity}.

\paragraph{Rank normalization and deterministic manifests.}
For each criterion \(m\), source example \(i\), and budget \(B\), let
\(\rho_{i,B}^{(m)}\in\{1,\ldots,N\}\) denote the rank assigned by criterion
\(m\), where rank 1 is best. TAVO converts ranks to a common-scale valuation
\begin{equation}
    v_{i,B}^{(m)}
    =
    \frac{N-\rho_{i,B}^{(m)}}{N-1},
    \qquad
    v_{i,B}^{(m)}\in[0,1].
\end{equation}
A fusion vector \(\mathbf{w}\in\Delta^{M-1}\) induces the fused score
\begin{equation}
    s_{i,B}(\mathbf{w})
    =
    \sum_{m=1}^{M} w_m v_{i,B}^{(m)}.
\end{equation}
The selected source manifest \(\mathcal{A}_{B}(\mathbf{w})\) contains the
top-\(B\) source cases by \(s_{i,B}(\mathbf{w})\), with ties resolved by fixed
source-identifier order. Thus, once criterion rankings are cached, TAVO is
deterministic for a given target, budget, and fusion vector.

\subsection{Warm-Up Model and Selection-Time Evidence}
\label{supp:warmup_protocol}

For each held-out target task, we train one Target + Full Source warm-up model
from the same task-specific initialization $\theta_{\mathrm{init}}$ used for
candidate and final models. Its frozen checkpoint is used only to compute
selection-time representations and gradients. The resulting cached evidence is
shared by all eight TAVO criteria, budgets, and fusion candidates for that
target. For TAVO, $T^\ell_{\mathrm{val}}$ is not used to construct the eight
criterion rankings or to train final models; it is reserved for outer-loop
fusion selection and checkpoint selection.

The evidence checkpoint is a fixed early checkpoint rather than a
test-selected checkpoint and is denoted \texttt{epoch005} in our experiments.
After source selection, every candidate proxy model and final reported model is
reinitialized from $\theta_{\mathrm{init}}$; no candidate or final model
continues training from the warm-up checkpoint.

\begin{table}[ht]
\centering
\caption{
Selection-time warm-up models. The frozen warm-up checkpoint is used only to
cache criterion evidence and is not used to initialize candidate or final
models.
}
\label{tab:supp_warmup_models}
\small
\setlength{\tabcolsep}{4pt}
\renewcommand{\arraystretch}{1.08}
\begin{adjustbox}{max width=\linewidth}
\begin{tabular}{llll}
\toprule
Task & Warm-up backbone & Warm-up data & Evidence unit \\
\midrule
BraTS & EfficientViT-Seg-L1 & Target + Full Source & patient case \\
MAMA-MIA & nnU-Net v2 2D & Target + Full Source & patient case \\
OfficeHome & ResNet-50 & Target + Full Source & image \\
\bottomrule
\end{tabular}
\end{adjustbox}
\end{table}

\begin{table}[ht]
\centering
\caption{
Task-specific warm-up configurations used to produce the frozen evidence
checkpoints. Hyperparameters follow the corresponding backbone training recipe
and are fixed across criteria and source budgets within each benchmark.
Final reported models are trained from scratch under the full-training schedules
in Sec.~\ref{app:implementation_details}.
}
\label{tab:supp_warmup_config}
\small
\setlength{\tabcolsep}{3.5pt}
\renewcommand{\arraystretch}{1.08}
\begin{adjustbox}{max width=\linewidth}
\begin{tabular}{lccc}
\toprule
Setting & BraTS & MAMA-MIA & OfficeHome \\
\midrule
Checkpoint used for evidence & epoch 5 & epoch 5 & epoch 5 \\
Optimizer & AdamW & SGD & AdamW \\
Learning rate & \(10^{-4}\) & \(10^{-2}\) & \(3\times10^{-4}\) \\
Batch size & 4 & 16 & 64 \\
Warm-up horizon & 5 epochs & 5 epochs & 5 epochs \\
Checkpoint rule & fixed epoch & fixed epoch & fixed epoch \\
\bottomrule
\end{tabular}
\end{adjustbox}
\end{table}

\subsection{Multi-Fidelity Search Configuration}
\label{supp:search_details}

TAVO performs validation-guided fusion search independently for each held-out
target domain and source budget. The search explicitly evaluates the \(M\)
simplex corners and the uniform fusion, then uses CMA-ES~\citep{hansen2001cmaes} to propose additional
fusion candidates. CMA-ES samples raw vectors
\(\mathbf{z}\in\mathbb{R}^{M}\), which are mapped to the probability simplex by
nonnegative normalization:
\begin{equation}
\operatorname{Norm}_{\Delta}(\mathbf{z})
=
\begin{cases}
[\mathbf{z}]_{+}/\|[\mathbf{z}]_{+}\|_{1},
& \|[\mathbf{z}]_{+}\|_{1}>0,\\
M^{-1}\mathbf{1},
& \text{otherwise},
\end{cases}
\end{equation}
where \([\cdot]_{+}\) denotes elementwise clipping at zero.

A candidate fusion vector is scored by proxy training on its induced source
subset and evaluating on the held-out target validation split. The coarse stage
uses one predetermined training seed, while the refinement stage reevaluates the
top candidates over three predetermined seeds and aggregates by the median.
Every refinement evaluation is restarted from \(\theta_{\mathrm{init}}\); it
does not continue from the corresponding coarse trajectory.

\begin{table}[ht]
\centering
\caption{
Multi-fidelity search configuration. These settings are fixed before final test
evaluation for each target--budget pair.
}
\label{tab:supp_search_configuration}
\small
\setlength{\tabcolsep}{4pt}
\renewcommand{\arraystretch}{1.08}
\begin{adjustbox}{max width=\linewidth}
\begin{tabular}{lc}
\toprule
Setting & Value \\
\midrule
Number of criteria \(M\) & 8 \\
Simplex corners evaluated & 8 \\
Uniform fusion evaluated & 1 \\
CMA-ES generations \(G\) & 12 \\
Population size \(\lambda\) & 20 \\
Refined candidates \(R\) & 8 \\
Coarse training seeds \(|\Xi_{\mathrm c}|\) & 1 \\
Refinement training seeds \(|\Xi_{\mathrm r}|\) & 3 \\
Seed aggregation & median \\
Coarse horizon \(J_{\mathrm c}\) & 500 iterations \\
Refinement horizon \(J_{\mathrm r}\) & 1500 iterations \\
Coarse checkpoint set \(\mathcal{K}_{J_{\mathrm c}}\) & final proxy checkpoint after 500 updates \\
Refinement checkpoint set \(\mathcal{K}_{J_{\mathrm r}}\) & final proxy checkpoint after 1500 updates\\
Initial CMA-ES scale \(\sigma_0\) & 0.30 \\
CMA-ES parent count \(\mu\) & 10 \\
Validation support for fitness & full \(\mathcal{T}_{\mathrm{val}}^{\ell}\) \\
Tie-breaking & fixed source-identifier order \\
\bottomrule
\end{tabular}
\end{adjustbox}
\end{table}

All coarse configurations are pooled, including simplex corners, the uniform
fusion, and CMA-ES candidates. The \(R\) configurations with the highest coarse
validation fitness are reevaluated at refinement fidelity. The final fusion
vector is the configuration with the highest median refined validation fitness.
The CMA-ES sampling seed is independent of the neural-network training seeds.

\subsection{Search Budget and Computational Cost}
\label{supp:computational_cost}

With \(M\) simplex corners, one uniform fusion, \(G\) generations, population
size \(\lambda\), and \(R\) refined candidates, the total number of candidate
network-training runs per target--budget search is
\begin{equation}
N_{\mathrm{runs}}
=
\left(M+1+G\lambda\right)|\Xi_{\mathrm c}|
+
R|\Xi_{\mathrm r}|.
\label{eq:supp_search_run_count}
\end{equation}
For the main configuration, \(M=8\), \(G=12\), \(\lambda=20\), \(R=8\),
\(|\Xi_{\mathrm c}|=1\), and \(|\Xi_{\mathrm r}|=3\). This corresponds to 249
coarse runs and 24 refinement runs, or 273 candidate-training runs per
target--budget search. These candidate runs use proxy-training horizons rather
than the final full-training horizon.

\begin{table}[ht]
\centering
\caption{
Measured GPU-hours for representative tumor-segmentation settings at source
budget \(B=150\) on an NVIDIA A40 GPU. Training costs are reported for one
final-training seed. Curation overhead is the sum of warm-up, evidence
extraction, criterion scoring, and multi-fidelity search.
}
\label{tab:supp_computational_cost_b150}
\small
\setlength{\tabcolsep}{3.0pt}
\renewcommand{\arraystretch}{1.02}
\begin{adjustbox}{max width=\linewidth}
\begin{tabular}{lcc}
\toprule
Cost component & BraTS & MAMA-MIA\\
\midrule
Criterion warm-up checkpoint & 7.8 & 6.9 \\
Selection evidence extraction & 5.6 & 4.1 \\
Criterion scoring & 0.2 & 0.2 \\
Multi-fidelity search & 18.5 & 15.9 \\
\midrule
Curation overhead & 32.1 & 27.1 \\
Final training & 6.0 & 5.7 \\
TAVO end-to-end & 38.1 & 32.8 \\
Target + Full Source training & 32.0 & 28.6 \\
\midrule
End-to-end / full-source & \(1.19\times\) & \(1.15\times\) \\
Curation overhead / final training & \(5.35\times\) & \(4.75\times\) \\
Full-fidelity eight-selector sweep & 61.6 & 56.8 \\
\bottomrule
\end{tabular}
\end{adjustbox}
\end{table}

The source budget constrains candidate and final training. The warm-up and
valuation stages access the full source pool once per held-out target to cache
source evidence. Cached evidence can be reused across budgets for the same
held-out target when the same warm-up checkpoint and feature/gradient blocks
are used. Once the final source manifest is fixed, TAVO adds no deployment- or
inference-time computation. Thus, its one-time search cost is confined to
development while optimizing source composition under a fixed cardinality
budget.
\section{Experimental Details}
\label{app:experimental_details}

This section reports the fixed split manifests, source-pool sizes,
implementation details, and metric definitions used by the experiments in the
main paper. All target test sets are accessed only for final reporting, and all
budgeted methods use the same labeled target training support and exactly the
specified source-case budget during candidate and final training.

\subsection{Dataset Splits and Source Budgets}
\label{app:split_details}
\paragraph{Split construction and leakage control.}
All split manifests were constructed once before source selection, fusion search, and final training, and the same manifests are shared by all methods, budgets, and final-training seeds. For BraTS 2021~\citep{baid2021brats,menze2015brats,bakas2017advancing}
and MAMA-MIA~\citep{garrucho2025mamamia}, splitting is performed at the patient-case level using a fixed split seed of 2026; no 2D slice from a patient case appears in more than one of the target training, validation, and test splits. Source pools exclude all cases from the held-out target domain. Case selection is performed before slice expansion: when a source case is selected, all retained slices from that case used by the downstream learner are included together. Any preprocessing-time slice filtering or foreground-slice retention rule is applied before selection and is shared by all methods. Domain/cohort labels are fixed by the dataset-provided center or cohort mappings. For OfficeHome~\citep{venkateswara2017deep}, target support is class-balanced, with three target-training images and two validation images per class, and all remaining images from the held-out target domain are reserved for testing. All released split manifests use anonymized identifiers and remain immutable across experiments.
For BraTS and MAMA-MIA, all target-tuned comparisons use the fixed
validation support listed in Tables~6--7, with target test cases reserved
exclusively for final reporting. We scope our conclusions to these evaluated
validation regimes and do not extrapolate robustness to substantially smaller
validation supports.

\begin{table*}[tp]
\centering
\caption{BraTS 2021 leave-one-domain-out split sizes and source-budget
percentages for the labeled cases retained in the experimental split manifests.
Target cases are split into labeled training support, validation support, and
held-out test cases. Source-budget percentages are computed relative to the
labeled source pool for each held-out target domain.}
\label{tab:brats_split_budget}
\small
\setlength{\tabcolsep}{5.0pt}
\renewcommand{\arraystretch}{1.08}
\begin{adjustbox}{max width=\textwidth}
\begin{tabular}{lccccccccc}
\toprule
Target
& Target train
& Target val
& Target test
& Target total
& Source pool
& $B=50$
& $B=150$
& $B=250$
& $B=500$ \\
\midrule
C4
& 10 & 5 & 15 & 30 & 1221
& 4.1\% & 12.3\% & 20.5\% & 41.0\% \\
C5
& 12 & 10 & 25 & 47 & 1204
& 4.2\% & 12.5\% & 20.8\% & 41.5\% \\
TCGA-LGG
& 15 & 10 & 40 & 65 & 1186
& 4.2\% & 12.6\% & 21.1\% & 42.2\% \\
TCGA-GBM
& 27 & 15 & 60 & 102 & 1149
& 4.4\% & 13.1\% & 21.8\% & 43.5\% \\
\bottomrule
\end{tabular}
\end{adjustbox}
\end{table*}

\begin{table*}[tp]
\centering
\caption{MAMA-MIA leave-one-domain-out split sizes and source-budget
percentages. Target support cases are split into training and validation
support; held-out test cases are used only for final reporting. Source-budget
percentages are computed relative to the labeled source pool for each held-out
target domain.}
\label{tab:mamamia_split_budget}
\small
\setlength{\tabcolsep}{5.0pt}
\renewcommand{\arraystretch}{1.08}
\begin{adjustbox}{max width=\textwidth}
\begin{tabular}{lccccccccc}
\toprule
Target
& Target train
& Target val
& Target test
& Target total
& Source pool
& $B=50$
& $B=150$
& $B=250$
& $B=500$ \\
\midrule
NACT
& 13 & 6 & 45 & 64 & 1442
& 3.5\% & 10.4\% & 17.3\% & 34.7\% \\
I-SPY1
& 34 & 17 & 120 & 171 & 1335
& 3.7\% & 11.2\% & 18.7\% & 37.5\% \\
Duke
& 58 & 25 & 208 & 291 & 1215
& 4.1\% & 12.3\% & 20.6\% & 41.2\% \\
I-SPY2
& 196 & 28 & 756 & 980 & 526
& 9.5\% & 28.5\% & 47.5\% & 95.1\% \\
\bottomrule
\end{tabular}
\end{adjustbox}
\end{table*}

\begin{table*}[tp]
\centering
\caption{OfficeHome multi-source target splits and source-budget percentages.
Each target domain uses three labeled target-training examples and two
validation examples per class. Source budgets are expressed as \(b\) selected
source images per class, so \(B=65b\). Source-budget percentages are computed
relative to the pooled source domains.}
\label{tab:officehome_split_budget}
\small
\setlength{\tabcolsep}{3.6pt}
\renewcommand{\arraystretch}{1.08}
\begin{adjustbox}{max width=\textwidth}
\begin{tabular}{lcccccccccccc}
\toprule
Target
& Target train
& Target val
& Target test
& Target total
& Source pool
& \(b=1\)
& \(b=3\)
& \(b=5\)
& \(b=8\)
& \(b=15\)
& \(b=25\)
& \(b=40\) \\
\midrule
Art
& 195 & 130 & 2102 & 2427 & 13161
& 0.5\% & 1.5\% & 2.5\% & 4.0\% & 7.4\% & 12.3\% & 19.8\% \\
Clipart
& 195 & 130 & 4040 & 4365 & 11223
& 0.6\% & 1.7\% & 2.9\% & 4.6\% & 8.7\% & 14.5\% & 23.2\% \\
Product
& 195 & 130 & 4114 & 4439 & 11149
& 0.6\% & 1.7\% & 2.9\% & 4.7\% & 8.7\% & 14.6\% & 23.3\% \\
Real World
& 195 & 130 & 4032 & 4357 & 11231
& 0.6\% & 1.7\% & 2.9\% & 4.6\% & 8.7\% & 14.5\% & 23.2\% \\
\bottomrule
\end{tabular}
\end{adjustbox}
\end{table*}
\subsection{Implementation Details}
\label{app:implementation_details}

\paragraph{Shared selection and training protocol.}
For each held-out target task, we train one Target + Full Source warm-up model
using the corresponding task-specific architecture and training protocol. This
warm-up model is used only for source valuation: for the eight TAVO criteria, all feature-based and gradient-based criteria
use representations or gradients cached from the same frozen checkpoint. After source selection, every reported
final model is trained from scratch using the selected source set together with
the labeled target training support. The target validation split is used for
TAVO fusion selection and for checkpoint selection, and the target test split
is used only for final reporting.

\paragraph{BraTS implementation.}
BraTS volumes are preprocessed offline using the nnU-Net preprocessing pipeline~\citep{isensee2021nnunet}
and exported as 2D axial slices. Each slice keeps the four MRI modality
channels, is min--max normalized to $[0,1]$, and is resized to
$512\times512$. We use EfficientViT-Seg-L1 with ADE20K-pretrained weights as
the segmentation backbone
\citep{cai2023efficientvit,zhou2017scene}, adapting the input layer to four
channels. Models are trained for 20 epochs with batch size 4 using AdamW,
learning rate $1\times10^{-4}$, weight decay $1\times10^{-5}$, cosine
annealing with $\eta_{\min}=1\times10^{-6}$, and a combined
Dice--cross-entropy loss. Final results are reported over training seeds
0, 1, and 2.

\paragraph{MAMA-MIA implementation.}
MAMA-MIA experiments use nnU-Net v2~\citep{isensee2021nnunet} with the 2D configuration and three DCE
input channels, denoted as \texttt{dce\_0000}, \texttt{dce\_0001}, and
\texttt{dce\_0002}. We use fold 0 for the nnU-Net fold split. Models are
trained for 100 epochs under the standard nnU-Net v2 training protocol. The
default optimizer is SGD with learning rate $1\times10^{-2}$, momentum
$0.99$, Nesterov acceleration enabled, and weight decay $3\times10^{-5}$.
The learning-rate schedule follows PolyLR. Each epoch uses 250 training
iterations and 50 validation iterations. Foreground oversampling is enabled
with oversampling fraction 0.33, and deep supervision is used. All MAMA-MIA
source-selection methods use the same nnU-Net preprocessing, fold split, and
final-training protocol.

\paragraph{Segmentation domain-adaptation baselines.}
DANN~\citep{ganin2016dann},
MMD~\citep{long2015dan},
ADVENT~\citep{vu2019advent}, and
SE-ASA~\citep{feng2023seasa} are trained with the same selected source budget
and labeled target training support as the budgeted source-selection methods.
For the budget-matched DA baselines in the main comparison, the selected source
set is a Random-$B$ subset. Thus, the DA baselines evaluate model-centric
adaptation under the same source-cardinality constraint, while TAVO evaluates
data-centric source composition under the standard segmentation objective.
All segmentation DA baselines use the same preprocessing, backbone, final
training horizon, target validation checkpoint-selection rule, and final test
protocol as the corresponding ERM baselines. The SE-ASA entries in the main
comparison correspond to the original configuration; the additional
validation-tuned SE-ASA analysis described below is reported separately.

\paragraph{Additional segmentation comparisons.}
At the representative source budget $B=150$, we additionally evaluate
GLISTER, validation-tuned SE-ASA, and a controlled BraTS nnU-Net comparison.
All experiments use the same fixed target splits and held-out test protocol
as the corresponding main experiments.

\textbf{GLISTER.}
We implement GLISTER~\citep{killamsetty2021glister} as a case-level,
target-aware selector using its one-step Taylor approximation and RGreedy
selection. Source cases are selected using target-validation utility, with
gradients computed from the segmentation head, and the resulting $B=150$
subset is trained under the same backbone, schedule, checkpoint rule, and
three final-training seeds as the other fixed selectors. Target-test cases
are never used during selection. Results are reported in Supplementary
Sec.~C.5.

\textbf{Validation-tuned SE-ASA.}
To provide a stronger tuned DA counterpart, we perform a compact
validation-based sensitivity sweep around the adapter-default SE-ASA
coefficients used in our main comparison. The original SE-ASA formulation~\citep{feng2023seasa}
uses fixed weights for its adversarial, prototype-alignment, and selective-
entropy objectives; because our EfficientViT and nnU-Net implementations
expose different adapter-specific parameterizations, we define separate
five-point local grids for BraTS and MAMA-MIA.

The grids probe the principal adaptation-loss trade-offs without an
exhaustive hyperparameter search: for BraTS, each non-default setting
perturbs one major loss weight, whereas the MAMA-MIA grid varies the overall
alignment strength and the relative emphasis of entropy versus semantic
alignment. For each held-out target, all configurations are evaluated on the
predefined target-validation split; the best validation configuration is
fixed before held-out test evaluation. Exact grids, selected configurations,
and test results are reported in Supplementary Sec.~C.5.

\textbf{Controlled BraTS nnU-Net comparison.}
We additionally repeat the BraTS $B=150$ comparison using nnU-Net v2~\citep{isensee2021nnunet}
with the self-configured \texttt{nnUNetPlans\_\_2d} setting and fold 0,
following the same nnU-Net v2 training framework used for MAMA-MIA and
adapting it to the four BraTS MRI modalities. All methods use the same fixed target splits and final-training protocol; the budgeted methods use the same $B=150$ source-cardinality constraint, while the full-source entries are reported as non-budgeted references. Complete Dice and HD95 results are reported in Supplementary Sec.~C.5.

\paragraph{OfficeHome implementation.}
For OfficeHome~\citep{venkateswara2017deep}, all classification models use an ImageNet-pretrained ResNet-50~\citep{he2016deep}
backbone. Models are trained for 30 epochs with batch size 64, learning rate
$3\times10^{-4}$, weight decay $1\times10^{-4}$, and cosine scheduling
with $T_{\max}$ equal to the number of training epochs. Training images are
resized to 256, randomly resized-cropped to 224, randomly horizontally flipped,
and normalized using ImageNet statistics. Evaluation images are resized to 256,
center-cropped to 224, and normalized with the same ImageNet statistics. Final
results are reported over training seeds 0, 1, and 2.

For classification DA baselines, the supervised objective uses cross-entropy on
both labeled source examples and labeled target-support examples. The
adaptation loss aligns the selected source training set with the labeled target
support set. We evaluate DANN~\citep{ganin2016dann},
MMD~\citep{long2015dan},
CORAL~\citep{sun2016coral}, and
CDAN~\citep{long2018cdan} under the same source-budget
protocol as the source-selection methods. The adaptation loss weight uses
$\lambda_{\max}=0.1$. For CDAN, we use entropy-conditioned CDAN
(\texttt{cdan\_e}) following the configuration used in our implementation.

\subsection{Metric Definitions}
\label{app:metric_definitions}

\paragraph{BraTS segmentation metrics.}
For BraTS, 2D slice predictions are stacked back into 3D volumes before
evaluation. The main paper reports 3D case-level Macro Dice averaged over ET,
TC, and WT as the primary BraTS metric. Dice is computed at the 3D case level for the enhancing
tumor (ET), tumor core (TC), and whole tumor (WT) regions. Macro Dice is obtained by averaging the corresponding ET, TC, and WT scores for each
case, and target-level results are then averaged over held-out test cases.

\paragraph{MAMA-MIA segmentation metrics.}
For MAMA-MIA, we report case-level foreground Dice as the primary metric for
the main results. Metrics are computed at the
case level on held-out target test cases and then averaged within each target
domain.

\paragraph{HD95 metrics.}
For the supplementary diagnostic surface-distance evaluation, HD95 is computed on reconstructed 3D volumes. For BraTS, HD95 is computed separately for ET, TC, and WT and then averaged to obtain Macro HD95 at the case level before target-level averaging. For MAMA-MIA, foreground HD95 is computed on the binary tumor mask. HD95 is reported only as a post-hoc diagnostic metric and is not used for TAVO fusion selection or checkpoint selection.

\paragraph{Classification metrics.}
For OfficeHome, the primary metric is target-test Accuracy. We additionally
support Macro-F1 and balanced accuracy for supplementary diagnostics. Macro-F1
is computed as the unweighted average of class-wise F1 scores over the 65
object categories, and balanced accuracy is computed as the unweighted average
of class-wise recall.

\section{Expanded Results}
\label{app:additional_results}

This section provides the expanded results referenced by the main paper.
Per-target/domain values are reported as mean$\pm$standard deviation over
three final-training seeds unless otherwise stated; dataset-level averages
are unweighted averages of the four target means. In addition
to the complete source-budget and OfficeHome results, we report expanded
ablations, paired target-level significance tests, additional strong
comparators, search-budget sensitivity, and criterion-ranking diagnostics.

\subsection{Complete Per-Target Results Across Source Budgets}
\label{app:per_target_budget_results}
This subsection reports the complete tumor-segmentation results across all
evaluated source budgets. \Cref{tab:supp_nonbudgeted_refs} summarizes the
non-budgeted references, while \Cref{tab:supp_brats_all_budget_results} and
\Cref{tab:supp_mamamia_all_budget_results} provide the budget-matched
per-target results for BraTS 2021 and MAMA-MIA, respectively.

\begin{table*}[tp]
\centering
\caption{Non-budgeted references for BraTS 2021 and MAMA-MIA.}
\label{tab:supp_nonbudgeted_refs}
\small
\setlength{\tabcolsep}{2.8pt}
\renewcommand{\arraystretch}{1.02}
\begin{adjustbox}{max width=\textwidth}
\begin{tabular}{lccccc@{\hspace{1.4em}}ccccc}
\toprule
& \multicolumn{5}{c}{BraTS 2021 Macro Dice \(\uparrow\)}
& \multicolumn{5}{c}{MAMA-MIA Foreground Dice \(\uparrow\)} \\
\cmidrule(lr){2-6}\cmidrule(lr){7-11}
Method
& C4 & C5 & LGG & GBM & Avg.
& NACT & I-SPY1 & Duke & I-SPY2 & Avg. \\
\midrule
Target-only
& \valstd{70.7}{0.3}
& \valstd{77.7}{0.1}
& \valstd{56.5}{0.3}
& \valstd{82.4}{0.2}
& 71.8
& \valstd{53.6}{0.7}
& \valstd{66.1}{0.5}
& \valstd{56.3}{0.1}
& \valstd{74.7}{0.2}
& 62.7 \\

Source-only Full
& \valstd{76.0}{0.6}
& \valstd{73.4}{0.2}
& \valstd{59.5}{0.7}
& \valstd{79.0}{0.4}
& 72.0
& \valstd{71.5}{0.4}
& \valstd{69.6}{0.3}
& \valstd{56.0}{0.4}
& \valstd{62.5}{0.8}
& 64.9 \\

Target + Full Source
& \valstd{76.1}{0.5}
& \valstd{79.1}{0.3}
& \valstd{61.6}{0.4}
& \valstd{83.2}{0.2}
& 75.0
& \valstd{72.1}{0.5}
& \valstd{70.6}{0.2}
& \valstd{59.4}{0.5}
& \valstd{75.4}{0.6}
& 69.4 \\
\bottomrule
\end{tabular}
\end{adjustbox}
\end{table*}

\begin{table*}[tp]
\centering
\caption{BraTS 2021 Macro Dice across source budgets.}
\label{tab:supp_brats_all_budget_results}
\small
\setlength{\tabcolsep}{2.3pt}
\renewcommand{\arraystretch}{0.96}
\begin{adjustbox}{max width=\textwidth}
\begin{tabular}{lccccc@{\hspace{1.4em}}ccccc}
\toprule
& \multicolumn{5}{c}{\(B=50\)}
& \multicolumn{5}{c}{\(B=150\)} \\
\cmidrule(lr){2-6}\cmidrule(lr){7-11}
Method & C4 & C5 & LGG & GBM & Avg.
& C4 & C5 & LGG & GBM & Avg. \\
\midrule
Random & \valstd{74.8}{0.3} & \valstd{81.6}{0.5} & \valstd{57.1}{0.4} & \valstd{83.6}{0.2} & 74.3
& \valstd{75.9}{1.1} & \valstd{82.4}{0.8} & \valstd{60.9}{0.6} & \valstd{83.4}{0.3} & 75.6 \\
RDS & \valstd{74.7}{0.1} & \valstd{82.0}{0.1} & \valstd{58.5}{0.3} & \valstd{84.1}{0.2} & 74.8
& \valstd{76.5}{0.1} & \valstd{81.4}{0.2} & \valstd{63.8}{1.8} & \valstd{84.6}{0.2} & 76.6 \\
LESS & \valstd{75.0}{0.3} & \valstd{80.3}{0.5} & \valstd{56.7}{0.7} & \valstd{82.6}{0.4} & 73.6
& \valstd{77.9}{1.0} & \valstd{80.8}{0.4} & \valstd{62.6}{0.6} & \valstd{83.8}{0.2} & 76.3 \\
ORIENT & \valstd{74.5}{0.1} & \valstd{81.0}{0.1} & \valstd{58.6}{0.1} & \valstd{83.7}{0.2} & 74.4
& \valstd{76.9}{1.3} & \valstd{82.7}{0.4} & \valstd{61.7}{0.4} & \valstd{84.4}{0.0} & 76.4 \\
GradMatch & \valstd{75.3}{0.5} & \valstd{82.1}{0.1} & \valstd{57.7}{0.5} & \valstd{83.9}{0.2} & 74.7
& \valstd{78.5}{1.7} & \valstd{82.3}{0.1} & \valstd{62.4}{1.9} & \valstd{84.5}{0.4} & 76.9 \\
CRAIG & \valstd{75.1}{0.4} & \valstd{81.4}{0.2} & \valstd{56.5}{0.4} & \valstd{83.6}{0.3} & 74.1
& \valstd{78.8}{1.3} & \valstd{82.2}{0.4} & \valstd{61.8}{2.0} & \valstd{84.2}{0.0} & 76.8 \\
KMeans & \valstd{74.9}{0.2} & \valstd{81.8}{0.1} & \valstd{57.0}{0.2} & \valstd{83.4}{0.2} & 74.3
& \valstd{77.5}{1.3} & \valstd{82.2}{0.3} & \valstd{58.0}{1.4} & \valstd{84.3}{0.1} & 75.5 \\
KCenter & \valstd{73.6}{0.2} & \valstd{81.5}{0.4} & \valstd{57.4}{0.5} & \valstd{83.4}{0.1} & 74.0
& \valstd{78.7}{1.1} & \valstd{82.4}{0.1} & \valstd{62.4}{0.6} & \valstd{84.0}{0.2} & 76.8 \\
Diversity & \valstd{74.1}{0.3} & \valstd{79.9}{0.1} & \valstd{57.1}{0.4} & \valstd{82.6}{0.3} & 73.4
& \valstd{79.0}{1.7} & \valstd{80.2}{0.1} & \valstd{57.5}{1.1} & \valstd{83.2}{0.2} & 75.0 \\
DANN & \valstd{75.1}{0.1} & \valstd{80.8}{0.5} & \valstd{57.1}{0.4} & \valstd{83.3}{0.1} & 74.1
& \valstd{78.3}{2.0} & \valstd{82.2}{0.0} & \valstd{61.0}{2.4} & \valstd{84.4}{0.3} & 76.5 \\
MMD & \valstd{74.0}{0.7} & \valstd{80.8}{0.1} & \valstd{55.5}{0.8} & \valstd{82.4}{0.2} & 73.2
& \valstd{78.2}{0.7} & \valstd{81.7}{0.9} & \valstd{60.7}{1.5} & \valstd{84.2}{0.4} & 76.2 \\
ADVENT & \valstd{75.0}{0.6} & \valstd{80.8}{0.1} & \valstd{56.8}{0.7} & \valstd{83.2}{0.3} & 74.0
& \valstd{77.6}{2.2} & \valstd{82.1}{0.0} & \valstd{61.2}{0.4} & \valstd{84.5}{0.1} & 76.4 \\
SE-ASA & \valstd{74.8}{0.1} & \valstd{81.2}{0.1} & \valstd{56.5}{0.5} & \valstd{83.1}{0.3} & 73.9
& \valstd{80.2}{0.6} & \valstd{82.6}{0.2} & \valstd{60.9}{0.7} & \valstd{83.9}{0.1} & 76.9 \\
TAVO & \valstd{75.4}{0.3} & \valstd{82.6}{0.2} & \valstd{59.2}{0.4} & \valstd{84.0}{0.1} & 75.3
& \valstd{79.9}{0.6} & \valstd{83.0}{0.5} & \valstd{63.7}{0.9} & \valstd{84.9}{0.1} & 77.9 \\
\midrule[0.6pt]
& \multicolumn{5}{c}{\(B=250\)}
& \multicolumn{5}{c}{\(B=500\)} \\
\cmidrule(lr){2-6}\cmidrule(lr){7-11}
Method & C4 & C5 & LGG & GBM & Avg.
& C4 & C5 & LGG & GBM & Avg. \\
\midrule
Random & \valstd{76.8}{1.4} & \valstd{82.7}{0.4} & \valstd{60.7}{0.7} & \valstd{84.5}{0.5} & 76.2
& \valstd{75.6}{0.5} & \valstd{82.2}{0.7} & \valstd{61.8}{0.2} & \valstd{84.9}{0.5} & 76.1 \\
RDS & \valstd{78.2}{2.0} & \valstd{82.9}{0.2} & \valstd{62.8}{1.0} & \valstd{84.5}{0.1} & 77.1
& \valstd{77.1}{2.4} & \valstd{83.1}{0.2} & \valstd{62.9}{1.0} & \valstd{85.2}{0.1} & 77.1 \\
LESS & \valstd{77.7}{2.6} & \valstd{81.5}{0.2} & \valstd{62.0}{1.7} & \valstd{84.6}{0.2} & 76.5
& \valstd{76.9}{4.0} & \valstd{82.0}{0.2} & \valstd{63.6}{1.7} & \valstd{85.2}{0.2} & 76.9 \\
ORIENT & \valstd{79.1}{1.5} & \valstd{82.3}{0.4} & \valstd{62.3}{0.8} & \valstd{84.8}{0.3} & 77.1
& \valstd{76.8}{2.8} & \valstd{83.1}{0.4} & \valstd{63.9}{0.8} & \valstd{85.1}{0.3} & 77.2 \\
GradMatch & \valstd{79.0}{1.4} & \valstd{83.0}{0.2} & \valstd{62.1}{0.6} & \valstd{84.3}{0.3} & 77.1
& \valstd{75.7}{3.0} & \valstd{82.9}{0.2} & \valstd{63.4}{0.6} & \valstd{85.0}{0.3} & 76.8 \\
CRAIG & \valstd{77.3}{0.9} & \valstd{82.7}{0.3} & \valstd{61.2}{0.1} & \valstd{84.7}{0.1} & 76.5
& \valstd{75.8}{2.6} & \valstd{82.9}{0.3} & \valstd{61.4}{0.1} & \valstd{84.9}{0.1} & 76.3 \\
KMeans & \valstd{75.9}{0.2} & \valstd{82.6}{0.5} & \valstd{59.4}{0.8} & \valstd{84.6}{0.1} & 75.6
& \valstd{76.7}{2.8} & \valstd{82.8}{0.5} & \valstd{62.5}{0.8} & \valstd{84.9}{0.1} & 76.7 \\
KCenter & \valstd{77.3}{0.5} & \valstd{82.4}{0.4} & \valstd{61.9}{1.0} & \valstd{84.4}{0.2} & 76.5
& \valstd{77.5}{2.2} & \valstd{83.0}{0.4} & \valstd{64.2}{1.0} & \valstd{84.9}{0.2} & 77.4 \\
Diversity & \valstd{76.1}{1.4} & \valstd{80.4}{0.2} & \valstd{56.7}{0.5} & \valstd{83.0}{0.2} & 74.0
& \valstd{76.2}{2.2} & \valstd{80.5}{0.2} & \valstd{56.6}{0.5} & \valstd{83.1}{0.2} & 74.1 \\
DANN & \valstd{79.1}{2.7} & \valstd{82.2}{0.2} & \valstd{60.2}{1.8} & \valstd{84.7}{0.2} & 76.6
& \valstd{78.9}{1.5} & \valstd{82.1}{0.4} & \valstd{60.4}{1.1} & \valstd{84.6}{0.3} & 76.5 \\
MMD & \valstd{78.0}{3.5} & \valstd{82.2}{0.5} & \valstd{60.4}{0.8} & \valstd{84.6}{0.2} & 76.3
& \valstd{77.8}{2.1} & \valstd{82.0}{0.3} & \valstd{60.8}{0.4} & \valstd{84.6}{0.2} & 76.3 \\
ADVENT & \valstd{78.2}{0.2} & \valstd{82.1}{0.2} & \valstd{61.1}{0.8} & \valstd{84.7}{0.1} & 76.5
& \valstd{78.0}{0.4} & \valstd{82.6}{0.2} & \valstd{61.0}{0.6} & \valstd{84.8}{0.2} & 76.6 \\
SE-ASA & \valstd{78.3}{0.5} & \valstd{82.3}{0.3} & \valstd{60.1}{1.0} & \valstd{84.5}{0.2} & 76.3
& \valstd{78.1}{0.5} & \valstd{82.5}{0.3} & \valstd{60.6}{1.2} & \valstd{84.9}{0.4} & 76.5 \\
TAVO & \valstd{79.0}{0.3} & \valstd{83.2}{0.2} & \valstd{63.1}{0.9} & \valstd{84.8}{0.1} & 77.5
& \valstd{79.6}{1.0} & \valstd{83.1}{0.4} & \valstd{64.2}{0.9} & \valstd{85.2}{0.1} & 78.0 \\
\bottomrule
\end{tabular}
\end{adjustbox}
\end{table*}

\begin{table*}[tp]
\centering
\caption{MAMA-MIA foreground Dice across source budgets.}
\label{tab:supp_mamamia_all_budget_results}
\small
\setlength{\tabcolsep}{2.3pt}
\renewcommand{\arraystretch}{0.96}
\begin{adjustbox}{max width=\textwidth}
\begin{tabular}{lccccc@{\hspace{1.4em}}ccccc}
\toprule
& \multicolumn{5}{c}{\(B=50\)}
& \multicolumn{5}{c}{\(B=150\)} \\
\cmidrule(lr){2-6}\cmidrule(lr){7-11}
Method & NACT & I-SPY1 & Duke & I-SPY2 & Avg.
& NACT & I-SPY1 & Duke & I-SPY2 & Avg. \\
\midrule
Random & \valstd{67.4}{1.2} & \valstd{67.6}{0.6} & \valstd{59.5}{0.3} & \valstd{75.5}{0.4} & 67.5
& \valstd{67.8}{0.5} & \valstd{69.2}{0.6} & \valstd{63.5}{0.3} & \valstd{76.3}{0.8} & 69.2 \\
RDS & \valstd{63.5}{0.5} & \valstd{66.9}{0.8} & \valstd{61.8}{0.6} & \valstd{75.4}{0.4} & 66.9
& \valstd{70.3}{0.8} & \valstd{65.8}{0.2} & \valstd{64.4}{0.4} & \valstd{76.4}{0.4} & 69.2 \\
LESS & \valstd{66.0}{0.3} & \valstd{69.0}{0.2} & \valstd{60.9}{0.2} & \valstd{74.4}{0.7} & 67.6
& \valstd{69.9}{0.3} & \valstd{67.8}{0.2} & \valstd{62.4}{0.3} & \valstd{75.1}{0.4} & 68.8 \\
ORIENT & \valstd{66.5}{0.4} & \valstd{70.2}{0.4} & \valstd{61.0}{0.4} & \valstd{76.0}{0.2} & 68.4
& \valstd{67.4}{0.3} & \valstd{71.1}{0.2} & \valstd{60.0}{0.7} & \valstd{76.2}{0.5} & 68.7 \\
GradMatch & \valstd{68.8}{0.3} & \valstd{68.5}{0.3} & \valstd{62.3}{0.1} & \valstd{76.6}{0.8} & 69.1
& \valstd{68.6}{0.5} & \valstd{68.9}{0.3} & \valstd{62.5}{0.5} & \valstd{76.2}{0.6} & 69.1 \\
CRAIG & \valstd{67.1}{0.7} & \valstd{66.9}{1.2} & \valstd{58.5}{0.8} & \valstd{76.4}{0.1} & 67.2
& \valstd{69.6}{0.2} & \valstd{67.7}{0.4} & \valstd{61.9}{0.4} & \valstd{75.7}{0.2} & 68.7 \\
KMeans & \valstd{63.1}{1.0} & \valstd{66.2}{0.6} & \valstd{57.4}{0.2} & \valstd{75.5}{0.1} & 65.5
& \valstd{66.0}{1.0} & \valstd{69.3}{0.4} & \valstd{62.4}{1.0} & \valstd{76.2}{0.5} & 68.5 \\
KCenter & \valstd{63.7}{0.6} & \valstd{68.0}{0.5} & \valstd{61.1}{0.6} & \valstd{75.4}{0.6} & 67.0
& \valstd{71.0}{0.2} & \valstd{68.1}{0.2} & \valstd{63.2}{0.5} & \valstd{75.8}{1.1} & 69.5 \\
Diversity & \valstd{67.7}{0.9} & \valstd{65.5}{0.3} & \valstd{60.2}{0.5} & \valstd{75.6}{0.3} & 67.2
& \valstd{65.9}{0.9} & \valstd{68.7}{0.3} & \valstd{60.6}{1.3} & \valstd{76.2}{0.3} & 67.8 \\
DANN & \valstd{66.0}{0.2} & \valstd{65.2}{0.6} & \valstd{58.8}{0.3} & \valstd{75.2}{0.6} & 66.3
& \valstd{67.0}{1.1} & \valstd{66.2}{0.1} & \valstd{61.6}{0.2} & \valstd{74.2}{0.6} & 67.3 \\
MMD & \valstd{64.3}{0.4} & \valstd{65.5}{0.4} & \valstd{59.1}{0.4} & \valstd{75.3}{0.2} & 66.1
& \valstd{68.1}{0.9} & \valstd{64.3}{0.3} & \valstd{62.7}{0.6} & \valstd{73.8}{0.4} & 67.2 \\
ADVENT & \valstd{65.7}{0.4} & \valstd{68.0}{0.9} & \valstd{60.3}{1.0} & \valstd{74.9}{1.0} & 67.2
& \valstd{67.5}{0.4} & \valstd{65.1}{0.2} & \valstd{62.9}{1.0} & \valstd{76.5}{0.4} & 68.0 \\
SE-ASA & \valstd{62.2}{0.7} & \valstd{67.3}{0.3} & \valstd{56.9}{0.7} & \valstd{77.0}{0.4} & 65.9
& \valstd{69.4}{0.3} & \valstd{68.9}{0.5} & \valstd{63.2}{0.6} & \valstd{75.4}{0.1} & 69.2 \\
TAVO & \valstd{68.5}{0.4} & \valstd{69.6}{0.5} & \valstd{62.8}{0.2} & \valstd{76.6}{0.5} & 69.4
& \valstd{72.9}{0.3} & \valstd{71.2}{0.2} & \valstd{64.2}{0.5} & \valstd{77.3}{0.1} & 71.4 \\
\midrule[0.6pt]
& \multicolumn{5}{c}{\(B=250\)}
& \multicolumn{5}{c}{\(B=500\)} \\
\cmidrule(lr){2-6}\cmidrule(lr){7-11}
Method & NACT & I-SPY1 & Duke & I-SPY2 & Avg.
& NACT & I-SPY1 & Duke & I-SPY2 & Avg. \\
\midrule
Random & \valstd{71.6}{0.5} & \valstd{70.9}{0.3} & \valstd{63.0}{0.2} & \valstd{72.4}{0.2} & 69.5
& \valstd{71.9}{0.1} & \valstd{72.3}{0.4} & \valstd{62.9}{0.6} & \valstd{73.4}{0.2} & 70.1 \\
RDS & \valstd{73.4}{1.0} & \valstd{75.1}{0.4} & \valstd{65.4}{1.1} & \valstd{73.1}{0.6} & 71.7
& \valstd{73.0}{0.6} & \valstd{75.2}{0.7} & \valstd{66.0}{0.4} & \valstd{73.8}{0.3} & 72.0 \\
LESS & \valstd{74.9}{0.3} & \valstd{75.0}{0.5} & \valstd{67.1}{0.7} & \valstd{74.2}{1.3} & 72.8
& \valstd{73.3}{0.4} & \valstd{76.0}{0.3} & \valstd{64.4}{0.5} & \valstd{74.6}{1.0} & 72.1 \\
ORIENT & \valstd{65.7}{0.3} & \valstd{74.9}{0.6} & \valstd{65.3}{0.5} & \valstd{73.2}{0.4} & 69.8
& \valstd{68.4}{0.2} & \valstd{76.1}{0.3} & \valstd{65.9}{1.3} & \valstd{75.0}{0.5} & 71.4 \\
GradMatch & \valstd{73.0}{0.4} & \valstd{75.7}{0.4} & \valstd{64.9}{0.4} & \valstd{73.1}{0.5} & 71.7
& \valstd{74.0}{0.3} & \valstd{73.7}{0.5} & \valstd{66.4}{0.7} & \valstd{72.9}{0.4} & 71.7 \\
CRAIG & \valstd{69.0}{0.3} & \valstd{73.7}{0.4} & \valstd{63.8}{0.5} & \valstd{73.7}{0.3} & 70.0
& \valstd{73.4}{0.8} & \valstd{74.2}{0.6} & \valstd{64.1}{0.7} & \valstd{73.8}{0.2} & 71.4 \\
KMeans & \valstd{71.4}{0.2} & \valstd{76.1}{0.7} & \valstd{65.2}{0.1} & \valstd{72.5}{0.9} & 71.3
& \valstd{72.5}{1.1} & \valstd{75.8}{0.6} & \valstd{65.3}{0.3} & \valstd{72.5}{0.4} & 71.5 \\
KCenter & \valstd{68.9}{0.6} & \valstd{75.8}{0.4} & \valstd{66.5}{0.4} & \valstd{72.3}{0.3} & 70.9
& \valstd{70.3}{0.6} & \valstd{75.2}{0.3} & \valstd{65.5}{0.6} & \valstd{72.9}{0.1} & 70.9 \\
Diversity & \valstd{68.2}{0.7} & \valstd{74.6}{0.2} & \valstd{65.0}{0.2} & \valstd{73.3}{0.1} & 70.3
& \valstd{69.7}{0.6} & \valstd{74.7}{0.8} & \valstd{64.1}{0.5} & \valstd{74.0}{0.8} & 70.6 \\
DANN & \valstd{74.8}{0.2} & \valstd{75.0}{0.5} & \valstd{64.1}{0.3} & \valstd{73.8}{0.7} & 71.9
& \valstd{73.2}{0.3} & \valstd{75.7}{0.6} & \valstd{62.4}{0.2} & \valstd{73.1}{1.2} & 71.1 \\
MMD & \valstd{75.2}{0.3} & \valstd{74.9}{0.2} & \valstd{63.8}{0.6} & \valstd{72.9}{0.2} & 71.7
& \valstd{73.5}{0.5} & \valstd{74.7}{0.2} & \valstd{65.3}{0.5} & \valstd{74.2}{0.6} & 71.9 \\
ADVENT & \valstd{73.2}{0.1} & \valstd{74.7}{0.4} & \valstd{64.9}{0.2} & \valstd{74.2}{0.4} & 71.7
& \valstd{74.2}{0.2} & \valstd{75.1}{0.9} & \valstd{64.2}{0.3} & \valstd{76.3}{0.5} & 72.5 \\
SE-ASA & \valstd{73.6}{0.6} & \valstd{73.5}{0.3} & \valstd{65.1}{0.8} & \valstd{73.6}{0.5} & 71.5
& \valstd{74.0}{0.4} & \valstd{73.4}{0.7} & \valstd{64.0}{0.5} & \valstd{72.4}{0.2} & 71.0 \\
TAVO & \valstd{75.6}{0.5} & \valstd{76.5}{0.3} & \valstd{67.5}{0.4} & \valstd{74.9}{0.4} & 73.7
& \valstd{74.8}{0.6} & \valstd{76.8}{0.2} & \valstd{66.3}{0.5} & \valstd{76.9}{0.4} & 73.7 \\
\bottomrule
\end{tabular}
\end{adjustbox}
\end{table*}

\subsection{OfficeHome Control Results}
\label{app:officehome_control_results}

OfficeHome is included as a compact cross-task control to evaluate whether the
same rank-fusion interface transfers beyond tumor segmentation. Source budgets
are class-balanced and specified by the number of selected source images per
class \(b\). \Cref{tab:supp_officehome_budget_sweep} reports the average
accuracy across the four held-out OfficeHome target domains for all evaluated
budgets. \Cref{tab:supp_officehome_b15_acc} reproduces the representative
\(b=15\) per-domain accuracy result reported in the main paper, and
\Cref{tab:supp_officehome_macro_f1_b15} reports the corresponding Macro-F1
diagnostic.

\begin{table*}[tp]
\centering
\caption{OfficeHome target-averaged accuracy across source budgets.}
\label{tab:supp_officehome_budget_sweep}
\small
\setlength{\tabcolsep}{3.0pt}
\renewcommand{\arraystretch}{1.02}
\begin{adjustbox}{max width=\textwidth}
\begin{tabular}{lccccccc}
\toprule
Method & \(b=1\) & \(b=3\) & \(b=5\) & \(b=8\) & \(b=15\) & \(b=25\) & \(b=40\) \\
\midrule
Random & 47.6 & 51.9 & 54.1 & 56.6 & 60.0 & 62.3 & 63.8 \\
KMeans & 50.9 & 54.8 & 56.2 & 58.1 & 60.7 & 62.2 & 64.1 \\
KCenter & 50.9 & 50.5 & 50.7 & 51.4 & 54.8 & 58.4 & 61.9 \\
Facility Location & 50.9 & 54.5 & 55.8 & 58.1 & 60.8 & 62.4 & 63.7 \\
GradMatch & 48.8 & 52.2 & 54.6 & 57.3 & 60.7 & 63.3 & 65.1 \\
CRAIG & 50.5 & 54.4 & 56.4 & 58.0 & 60.5 & 62.0 & 63.4 \\
GLISTER & 50.2 & 54.6 & 56.8 & 59.2 & 61.7 & 63.7 & 65.1 \\
ORIENT & 49.8 & 54.6 & 56.4 & 58.8 & 62.2 & 63.8 & 64.8 \\
MomentMatching & 49.0 & 52.2 & 54.2 & 57.7 & 60.6 & 63.3 & 65.1 \\
\midrule
DANN & 47.1 & 52.1 & 55.1 & 56.9 & 59.8 & 61.5 & 63.3 \\
MMD & 47.2 & 52.0 & 55.0 & 56.6 & 59.8 & 61.3 & 63.2 \\
CORAL & 47.2 & 51.9 & 55.2 & 56.7 & 59.6 & 61.5 & 63.2 \\
CDAN & 47.1 & 51.9 & 55.0 & 57.0 & 60.1 & 61.9 & 63.4 \\
\midrule
TAVO & 50.5 & 55.8 & 57.9 & 59.8 & 62.7 & 64.9 & 65.9 \\
\bottomrule
\end{tabular}
\end{adjustbox}
\end{table*}

\begin{table}[ht]
\centering
\caption{OfficeHome accuracy at \(b=15\) selected source images per class.}
\label{tab:supp_officehome_b15_acc}
\small
\setlength{\tabcolsep}{1.8pt}
\renewcommand{\arraystretch}{1.0}
\begin{adjustbox}{max width=\linewidth}
\begin{tabular}{lccccc}
\toprule
Method & Art & Clipart & Product & Real & Avg. \\
\midrule
Target-only & \valstd{43.9}{1.5} & \valstd{26.3}{0.4} & \valstd{56.6}{1.8} & \valstd{51.2}{1.6} & 44.5 \\
Source-only Full & \valstd{60.3}{1.1} & \valstd{40.3}{0.8} & \valstd{75.2}{0.2} & \valstd{75.5}{0.7} & 62.8 \\
Target + Full Source & \valstd{63.3}{1.0} & \valstd{49.8}{0.4} & \valstd{78.7}{0.5} & \valstd{77.0}{0.5} & 67.2 \\
\midrule
Random & \valstd{56.6}{1.3} & \valstd{44.0}{0.9} & \valstd{71.1}{0.3} & \valstd{68.3}{1.0} & 60.0 \\
\midrule
KMeans & \valstd{57.7}{0.3} & \valstd{44.7}{1.1} & \valstd{72.3}{1.5} & \valstd{68.0}{0.5} & 60.7 \\
KCenter & \valstd{49.5}{0.7} & \valstd{40.2}{0.2} & \valstd{67.7}{1.0} & \valstd{61.6}{1.6} & 54.7 \\
Facility Location & \valstd{58.0}{1.2} & \valstd{43.9}{0.8} & \valstd{72.4}{0.8} & \valstd{68.9}{0.5} & 60.8 \\
GradMatch & \valstd{58.6}{1.2} & \valstd{44.3}{0.9} & \valstd{73.6}{0.6} & \valstd{66.4}{1.3} & 60.7 \\
CRAIG & \valstd{55.4}{0.5} & \valstd{44.6}{0.3} & \valstd{72.7}{0.7} & \valstd{69.2}{0.4} & 60.5 \\
GLISTER & \valstd{59.7}{1.0} & \valstd{43.9}{1.3} & \valstd{74.9}{0.7} & \valstd{68.1}{0.7} & 61.7 \\
ORIENT & \valstd{60.6}{0.4} & \valstd{46.1}{1.2} & \valstd{73.5}{0.4} & \valstd{68.4}{0.4} & 62.2 \\
MomentMatching & \valstd{58.2}{1.1} & \valstd{44.1}{0.6} & \valstd{73.4}{0.1} & \valstd{66.8}{0.9} & 60.6 \\
\midrule
DANN & \valstd{57.9}{0.2} & \valstd{41.6}{0.3} & \valstd{70.9}{0.9} & \valstd{68.9}{0.8} & 59.8 \\
MMD & \valstd{58.5}{0.6} & \valstd{41.2}{0.7} & \valstd{70.6}{0.5} & \valstd{68.7}{0.4} & 59.8 \\
CORAL & \valstd{57.1}{0.3} & \valstd{41.6}{0.6} & \valstd{70.7}{0.2} & \valstd{68.9}{0.6} & 59.6 \\
CDAN & \valstd{58.3}{0.5} & \valstd{42.0}{0.5} & \valstd{70.9}{1.1} & \valstd{69.1}{0.4} & 60.1 \\
\midrule
TAVO & \valstd{59.7}{1.0} & \valstd{46.5}{0.3} & \valstd{75.2}{0.7} & \valstd{69.5}{0.7} & 62.7 \\
\bottomrule
\end{tabular}
\end{adjustbox}
\end{table}

\begin{table}[ht]
\centering
\caption{OfficeHome Macro-F1 at \(b=15\). Higher is better.}
\label{tab:supp_officehome_macro_f1_b15}
\small
\setlength{\tabcolsep}{1.8pt}
\renewcommand{\arraystretch}{1.0}
\begin{adjustbox}{max width=\linewidth}
\begin{tabular}{lccccc}
\toprule
Method & Art & Clipart & Product & Real & Avg. \\
\midrule
Random
& \valstd{51.2}{1.4}
& \valstd{43.1}{0.8}
& \valstd{69.4}{0.6}
& \valstd{65.9}{1.1}
& \valstd{57.4}{0.2} \\

KMeans
& \valstd{52.3}{0.4}
& \valstd{43.7}{1.0}
& \valstd{70.8}{1.4}
& \valstd{65.6}{0.5}
& \valstd{58.1}{0.2} \\

KCenter
& \valstd{43.7}{0.4}
& \valstd{39.1}{0.5}
& \valstd{65.5}{1.2}
& \valstd{58.5}{1.7}
& \valstd{51.7}{0.7} \\

Facility Location
& \valstd{52.4}{1.1}
& \valstd{42.7}{0.8}
& \valstd{70.3}{0.9}
& \valstd{66.3}{0.5}
& \valstd{57.9}{0.6} \\

CRAIG
& \valstd{50.7}{0.4}
& \valstd{43.5}{0.4}
& \valstd{71.1}{0.7}
& \valstd{67.0}{0.3}
& \valstd{58.1}{0.2} \\

MomentMatching
& \valstd{52.6}{1.1}
& \valstd{43.2}{0.8}
& \valstd{72.1}{0.3}
& \valstd{64.8}{0.8}
& \valstd{58.2}{0.3} \\

GradMatch
& \valstd{52.9}{1.0}
& \valstd{43.1}{1.0}
& \valstd{72.1}{0.7}
& \valstd{64.4}{1.2}
& \valstd{58.1}{0.1} \\

GLISTER
& \valstd{54.0}{0.7}
& \valstd{42.6}{1.1}
& \valstd{73.5}{0.5}
& \valstd{65.6}{0.7}
& \valstd{58.9}{0.3} \\

ORIENT
& \valstd{55.2}{0.8}
& \valstd{45.1}{1.3}
& \valstd{72.0}{0.4}
& \valstd{66.0}{0.5}
& \valstd{59.6}{0.1} \\

\midrule
DANN
& \valstd{51.8}{0.6}
& \valstd{40.6}{0.3}
& \valstd{68.9}{1.2}
& \valstd{66.6}{0.9}
& \valstd{57.0}{0.6} \\

MMD
& \valstd{52.7}{0.9}
& \valstd{40.2}{0.3}
& \valstd{68.7}{0.7}
& \valstd{66.3}{0.4}
& \valstd{57.0}{0.4} \\

CORAL
& \valstd{50.8}{0.7}
& \valstd{40.6}{0.5}
& \valstd{68.6}{0.1}
& \valstd{66.6}{0.7}
& \valstd{56.6}{0.1} \\

CDAN
& \valstd{52.4}{0.2}
& \valstd{40.8}{0.5}
& \valstd{69.0}{1.3}
& \valstd{66.7}{0.4}
& \valstd{57.2}{0.6} \\

\midrule
TAVO
& \valstd{53.1}{1.1}
& \valstd{44.0}{0.5}
& \valstd{73.5}{0.1}
& \valstd{66.9}{0.6}
& \valstd{59.4}{0.4} \\
\bottomrule
\end{tabular}
\end{adjustbox}
\end{table}

\subsection{Complete Per-Target Ablations}
\label{app:per_target_ablations}
This subsection expands the main-paper ablation study to all held-out tumor
segmentation targets. \Cref{tab:supp_ablation_per_target_combined} reports
the per-target results for source-structural-only fusion,
target-conditioned-only fusion, Uniform fusion, Val.-selected single,
Dirichlet random search, Shuffled-ranking CMA-ES, and Full TAVO.

\begin{table*}[tp]
\centering
\caption{Per-target ablation results at \(B=150\).}
\label{tab:supp_ablation_per_target_combined}
\small
\setlength{\tabcolsep}{2.6pt}
\renewcommand{\arraystretch}{1.02}
\begin{adjustbox}{max width=\textwidth}
\begin{tabular}{lccccc@{\hspace{1.5em}}ccccc}
\toprule
& \multicolumn{5}{c}{BraTS Macro Dice \(\uparrow\)}
& \multicolumn{5}{c}{MAMA-MIA Foreground Dice \(\uparrow\)} \\
\cmidrule(lr){2-6}\cmidrule(lr){7-11}
Variant & C4 & C5 & LGG & GBM & Avg.
& NACT & I-SPY1 & Duke & I-SPY2 & Avg. \\
\midrule
Source-structural only
& \valstd{75.8}{0.3}
& \valstd{81.6}{0.4}
& \valstd{63.3}{0.9}
& \valstd{84.1}{0.1}
& 76.2
& \valstd{69.7}{0.1}
& \valstd{67.6}{0.5}
& \valstd{63.5}{0.1}
& \valstd{76.1}{0.2}
& 69.2 \\

Target-conditioned only
& \valstd{77.1}{0.4}
& \valstd{82.4}{0.2}
& \valstd{62.6}{0.4}
& \valstd{83.2}{0.2}
& 76.3
& \valstd{71.2}{0.5}
& \valstd{70.1}{0.3}
& \valstd{64.2}{0.7}
& \valstd{75.7}{0.5}
& 70.3 \\

Uniform fusion
& \valstd{76.2}{0.2}
& \valstd{82.1}{0.1}
& \valstd{62.0}{0.8}
& \valstd{84.2}{0.4}
& 76.1
& \valstd{69.1}{0.4}
& \valstd{68.2}{0.3}
& \valstd{63.7}{0.4}
& \valstd{76.4}{0.3}
& 69.4 \\

Val.-selected single
& \valstd{76.9}{1.3}
& \valstd{80.2}{0.1}
& \valstd{62.4}{1.9}
& \valstd{84.2}{0.0}
& 75.9
& \valstd{67.8}{0.5}
& \valstd{68.9}{0.3}
& \valstd{61.9}{0.4}
& \valstd{75.7}{0.2}
& 68.5 \\

Dirichlet random search
& \valstd{75.0}{0.6}
& \valstd{81.8}{0.5}
& \valstd{60.2}{0.4}
& \valstd{83.8}{0.2}
& 75.2
& \valstd{67.9}{0.6}
& \valstd{67.4}{0.2}
& \valstd{63.0}{0.6}
& \valstd{73.4}{0.5}
& 67.9 \\

Shuffled-ranking CMA-ES
& \valstd{75.8}{0.4}
& \valstd{82.4}{0.2}
& \valstd{61.4}{0.6}
& \valstd{84.3}{0.1}
& 76.0
& \valstd{68.9}{0.2}
& \valstd{69.4}{0.5}
& \valstd{63.5}{0.7}
& \valstd{76.4}{0.4}
& 69.5 \\

Full TAVO
& \valstd{79.9}{0.6}
& \valstd{83.0}{0.5}
& \valstd{63.7}{0.9}
& \valstd{84.9}{0.1}
& 77.9
& \valstd{72.9}{0.3}
& \valstd{71.2}{0.2}
& \valstd{64.2}{0.5}
& \valstd{77.3}{0.1}
& 71.4 \\
\bottomrule
\end{tabular}
\end{adjustbox}
\end{table*}

\subsection{Paired Target-Level Significance}
\label{app:paired_significance}

At $B=150$, we evaluate paired target-level differences between TAVO and
the headline comparison methods. For each held-out target, performance is
first averaged over the three final-training seeds, after which the
TAVO--baseline difference is formed. We report the mean over the four
target-wise differences, a 95\% confidence interval, and a two-sided
one-sample $t$-test against zero. This is therefore a target-level paired
analysis after three-seed aggregation.

\begin{table}[t]
\centering
\caption{
Paired target-level comparisons at $B=150$. Mean gain denotes
TAVO minus the baseline in Dice points.
}
\label{tab:supp_paired_significance}
\small
\setlength{\tabcolsep}{3.2pt}
\renewcommand{\arraystretch}{1.05}
\begin{adjustbox}{max width=\linewidth}
\begin{tabular}{llccc}
\toprule
Benchmark & Comparison & Gain & 95\% CI & $p$ \\
\midrule
BraTS & TAVO--GradMatch & +0.95 & $[+0.19,+1.71]$ & .029 \\
BraTS & TAVO--SE-ASA    & +0.98 & $[-1.14,+3.09]$ & .238 \\
MAMA-MIA & TAVO--KCenter & +1.88 & $[+0.45,+3.30]$ & .025 \\
MAMA-MIA & TAVO--SE-ASA  & +2.18 & $[+0.52,+3.83]$ & .025 \\
\bottomrule
\end{tabular}
\end{adjustbox}
\end{table}

\Cref{tab:supp_paired_significance} shows that TAVO significantly improves
over the strongest fixed selector on both medical benchmarks. Against the
original SE-ASA baseline, the gain is significant on MAMA-MIA but not on
BraTS. The separately validation-tuned SE-ASA analysis is reported in
Sec.~\ref{app:strong_comparators}.

\subsection{Additional Strong-Comparator Analyses}
\label{app:strong_comparators}
\paragraph{GLISTER.}
We evaluate GLISTER at $B=150$ under the matched protocol described in
Sec.~\ref{app:implementation_details}. \Cref{tab:supp_glister} reports the
three-seed per-target results.

\begin{table}[t]
\centering
\caption{GLISTER segmentation results at $B=150$.}
\label{tab:supp_glister}
\small
\setlength{\tabcolsep}{3.2pt}
\renewcommand{\arraystretch}{1.05}
\begin{adjustbox}{max width=\linewidth}
\begin{tabular}{lccccc}
\toprule
Benchmark & 1 & 2 & 3 & 4 & Avg. \\
\midrule
BraTS
& $77.0\pm0.6$
& $82.0\pm0.4$
& $62.3\pm0.7$
& $83.3\pm0.2$
& 76.1 \\
MAMA-MIA
& $65.2\pm0.5$
& $67.5\pm0.3$
& $60.2\pm0.4$
& $75.9\pm0.4$
& 67.2 \\
\bottomrule
\end{tabular}
\end{adjustbox}
\end{table}

For BraTS, columns 1--4 denote C4, C5, TCGA-LGG, and TCGA-GBM;
for MAMA-MIA, they denote NACT, I-SPY1, Duke, and I-SPY2.
GLISTER reaches 76.1/67.2 average Dice on BraTS/MAMA-MIA,
compared with 77.9/71.4 for TAVO.

\paragraph{Validation-tuned SE-ASA.}
To assess whether the main comparison is sensitive to SE-ASA loss weighting,
we perform the compact validation-based local sweep described in
Sec.~\ref{app:implementation_details}. The sweep is centered on the
adapter-default configuration used in the main comparison with dataset-specific grids reflecting the corresponding adapter parameterizations. For BraTS, the non-default settings
perturb one principal adaptation loss at a time; for MAMA-MIA, the two-weight
adapter probes lower/higher overall alignment strength and relative
entropy-versus-semantic emphasis. The exact configurations are given in
\Cref{tab:supp_seasa_grids}. Configuration selection uses target-validation
Dice only, and the selected setting is fixed before held-out test evaluation.

\begin{table}[t]
\centering
\caption{
Compact dataset-specific SE-ASA validation-tuning grids centered on the
adapter-default loss weights. Dashes denote parameters not exposed by the
corresponding adapter.
}
\label{tab:supp_seasa_grids}
\scriptsize
\setlength{\tabcolsep}{3.0pt}
\renewcommand{\arraystretch}{1.04}
\begin{adjustbox}{max width=\linewidth}
\begin{tabular}{llccccc}
\toprule
Dataset & Variant
& $\lambda_{\rm adv}$
& $\lambda_{\rm cls}$
& $\lambda_{\rm sel}$
& $\lambda_{\rm ent}$
& $\lambda_{\rm sem}$ \\
\midrule
BraTS & \texttt{base}             & 0.003 & 0.10 & 0.01 & -- & -- \\
BraTS & \texttt{adv\_low}         & 0.001 & 0.10 & 0.01 & -- & -- \\
BraTS & \texttt{adv\_high}        & 0.010 & 0.10 & 0.01 & -- & -- \\
BraTS & \texttt{class\_low}       & 0.003 & 0.03 & 0.01 & -- & -- \\
BraTS & \texttt{selective\_high}  & 0.003 & 0.10 & 0.03 & -- & -- \\
\midrule
MAMA-MIA & \texttt{base}              & -- & -- & -- & 0.10 & 0.10 \\
MAMA-MIA & \texttt{both\_low}         & -- & -- & -- & 0.03 & 0.03 \\
MAMA-MIA & \texttt{both\_high}        & -- & -- & -- & 0.30 & 0.30 \\
MAMA-MIA & \texttt{entropy\_dominant} & -- & -- & -- & 0.10 & 0.03 \\
MAMA-MIA & \texttt{semantic\_dominant}& -- & -- & -- & 0.03 & 0.10 \\
\bottomrule
\end{tabular}
\end{adjustbox}
\end{table}

\begin{table}[t]
\centering
\caption{
Validation-selected SE-ASA configurations at $B=150$.
Val. Dice denotes the target-validation score used for configuration
selection. Per-target Test Dice is reported as mean$\pm$standard deviation
over three final-training seeds; Avg. is the unweighted average of the four
target means. All values are in percentage points.
}
\label{tab:supp_seasa_selected}
\small
\setlength{\tabcolsep}{3.0pt}
\renewcommand{\arraystretch}{1.04}
\begin{adjustbox}{max width=\linewidth}
\begin{tabular}{llccc}
\toprule
Dataset & Target & Selected & Val. Dice & Test Dice \\
\midrule
BraTS & C4
& \texttt{base}
& 70.1 & $78.3 \pm 0.03$ \\

BraTS & C5
& \texttt{adv\_low}
& 80.7 & $81.9 \pm 0.20$ \\

BraTS & LGG
& \texttt{adv\_high}
& 65.9 & $61.2 \pm 1.26$ \\

BraTS & GBM
& \texttt{class\_low}
& 83.8 & $84.1 \pm 0.14$ \\

BraTS & Avg.
& --
& 75.1 & $76.4$ \\
\midrule

MAMA-MIA & NACT
& \texttt{base}
& 73.1 & $70.0 \pm 0.08$ \\

MAMA-MIA & I-SPY1
& \texttt{both\_high}
& 75.0 & $70.3 \pm 0.25$ \\

MAMA-MIA & Duke
& \texttt{both\_high}
& 70.8 & $63.4 \pm 0.22$ \\

MAMA-MIA & I-SPY2
& \texttt{entropy\_dominant}
& 78.9 & $76.2 \pm 0.38$ \\

MAMA-MIA & Avg.
& --
& 74.5 & $70.0$ \\
\bottomrule
\end{tabular}
\end{adjustbox}
\end{table}

The validation-selected SE-ASA configurations are then evaluated over 3
final-training seeds, obtaining target-averaged test Dice of
$76.4$ on BraTS and $70.0$ on MAMA-MIA.
For reference, the corresponding three-seed TAVO averages are
77.9 and 71.4, respectively.

\paragraph{Controlled BraTS nnU-Net comparison.}
Under the controlled nnU-Net setting described in
Sec.~\ref{app:implementation_details}, all methods use the same fixed target splits and final-training protocol;
the budgeted methods use the same $B=150$ source-cardinality constraint,
while the full-source entries are reported as non-budgeted references.

\begin{table*}[t]
\centering
\caption{
Controlled BraTS nnU-Net comparison at $B=150$.
Dice is higher-is-better; HD95 is lower-is-better.
}
\label{tab:supp_nnunet_control}
\scriptsize
\setlength{\tabcolsep}{2.3pt}
\renewcommand{\arraystretch}{1.03}
\begin{adjustbox}{max width=\textwidth}
\begin{tabular}{lccccc|ccccc}
\toprule
& \multicolumn{5}{c|}{Macro Dice $\uparrow$}
& \multicolumn{5}{c}{Macro HD95 $\downarrow$} \\
Method
& C4 & C5 & LGG & GBM & Avg.
& C4 & C5 & LGG & GBM & Avg. \\
\midrule
Target-only
& $76.5\pm0.3$ & $84.0\pm0.3$ & $58.5\pm0.4$ & $86.5\pm0.5$ & 76.4
& $22.841\pm4.4$ & $8.738\pm2.1$ & $51.278\pm3.2$ & $11.743\pm1.6$ & 23.650 \\

Source-only Full
& $81.0\pm0.5$ & $\mathbf{88.5\pm0.3}$ & $66.0\pm0.7$ & $89.3\pm0.2$ & 81.2
& $24.124\pm5.8$ & $2.928\pm5.3$ & $\mathbf{29.226\pm3.9}$ & $5.147\pm2.8$ & 15.356 \\

Target + Full Source
& $83.6\pm0.6$ & $86.2\pm0.2$ & $67.3\pm0.7$ & $88.9\pm0.4$ & 81.5
& $\mathbf{11.624\pm6.1}$ & $3.320\pm4.9$ & $31.314\pm7.1$ & $6.993\pm4.4$ & 13.313 \\

GradMatch-$B150$
& $79.6\pm0.4$ & $86.7\pm0.4$ & $66.7\pm1.0$ & $89.2\pm0.1$ & 80.5
& $22.931\pm8.5$ & $3.195\pm3.6$ & $35.826\pm3.8$ & $6.481\pm3.4$ & 17.108 \\

SE-ASA-$B150$
& $81.8\pm0.2$ & $86.9\pm0.3$ & $64.9\pm0.5$ & $88.6\pm0.2$ & 80.5
& $19.584\pm4.7$ & $3.118\pm4.4$ & $40.019\pm6.5$ & $5.964\pm2.8$ & 17.171 \\

TAVO-$B150$
& $\mathbf{84.3\pm0.3}$ & $88.4\pm0.4$ & $\mathbf{67.6\pm0.8}$ &
$\mathbf{89.6\pm0.1}$ & $\mathbf{82.5}$
& $13.384\pm5.3$ & $\mathbf{2.790\pm2.7}$ & $31.507\pm4.5$ &
$\mathbf{5.075\pm1.9}$ & $\mathbf{13.189}$ \\
\bottomrule
\end{tabular}
\end{adjustbox}
\end{table*}

TAVO achieves the highest average Dice (82.5) and lowest average HD95
(13.189) within this controlled nnU-Net comparison. These values are not
directly compared numerically with the EfficientViT-based BraTS results in
the main table because they use a different downstream backbone.

\subsection{Search-Budget Sensitivity}
\label{app:search_budget}

We evaluate the sensitivity of TAVO to fusion-search computation at
$B=150$, while keeping all other search and final-training settings fixed
as in Sec.~A.3. Reduced-budget settings use prefixes of the same full
CMA-ES trajectory rather than independent restarts. The 25/50/75/100\%
budgets correspond to 3/6/9/12 CMA-ES generations,
69/129/189/249 coarse evaluations, and 2/4/6/8 refined candidates,
respectively. All settings retain the same eight
simplex corners, uniform initialization, and CMA-ES population size used in the full search.

\begin{table*}[tp]
\centering
\caption{
Search-budget sensitivity at $B=150$. Values are mean$\pm$standard deviation
over three final-training seeds and are reported in percentage points.
Dataset averages are computed over the four target domains.
}
\label{tab:supp_search_budget}
\small
\setlength{\tabcolsep}{2.6pt}
\renewcommand{\arraystretch}{1.02}
\begin{adjustbox}{max width=\textwidth}
\begin{tabular}{lccccc@{\hspace{1.5em}}ccccc}
\toprule
& \multicolumn{5}{c}{BraTS Macro Dice $\uparrow$}
& \multicolumn{5}{c}{MAMA-MIA Foreground Dice $\uparrow$} \\
\cmidrule(lr){2-6}\cmidrule(lr){7-11}
Search
& C4 & C5 & LGG & GBM & Avg.
& NACT & I-SPY1 & Duke & I-SPY2 & Avg. \\
\midrule

25\%
& \valstd{76.3}{1.3}
& \valstd{82.6}{0.4}
& \valstd{61.9}{0.7}
& \valstd{84.1}{0.5}
& 76.2
& \valstd{68.5}{0.6}
& \valstd{69.7}{0.6}
& \valstd{61.1}{0.3}
& \valstd{75.5}{0.4}
& 68.7 \\

50\%
& \valstd{77.8}{0.7}
& \valstd{82.3}{0.2}
& \valstd{62.3}{0.6}
& \valstd{84.1}{0.1}
& 76.6
& \valstd{69.9}{0.4}
& \valstd{70.0}{0.2}
& \valstd{61.6}{0.8}
& \valstd{75.2}{0.3}
& 69.2 \\

75\%
& \valstd{78.3}{1.2}
& \valstd{82.8}{0.5}
& \valstd{64.1}{0.3}
& \valstd{84.4}{0.8}
& 77.4
& \valstd{69.8}{0.4}
& \valstd{69.8}{0.3}
& \valstd{63.5}{0.7}
& \valstd{75.6}{0.1}
& 69.7 \\

100\%
& \valstd{79.9}{0.6}
& \valstd{83.0}{0.5}
& \valstd{63.7}{0.9}
& \valstd{84.9}{0.1}
& 77.9
& \valstd{72.9}{0.3}
& \valstd{71.2}{0.2}
& \valstd{64.2}{0.5}
& \valstd{77.3}{0.1}
& 71.4 \\

\bottomrule
\end{tabular}
\end{adjustbox}
\end{table*}

\begin{figure}[!t]
  \centering
  \includegraphics[width=\linewidth]
  {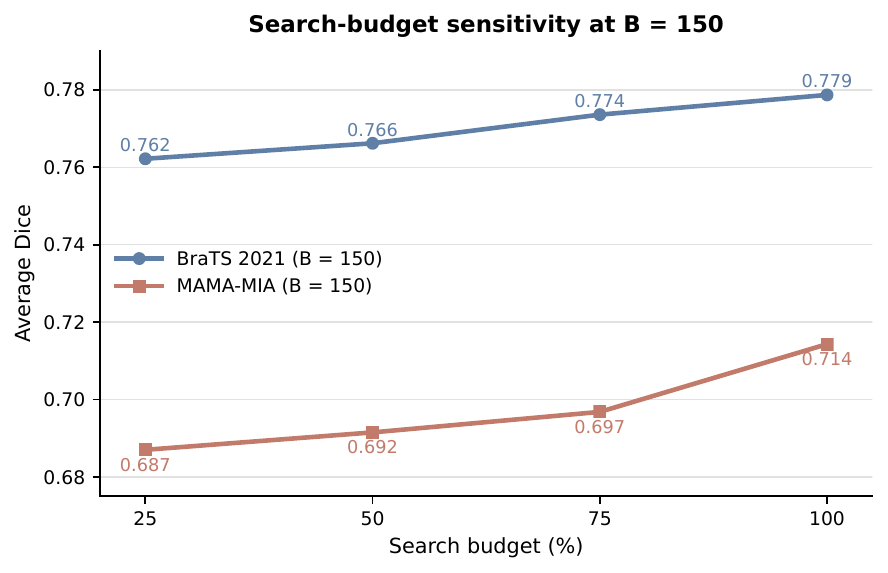}
  \caption{
  Target-averaged Dice versus fusion-search budget at $B=150$.
  Reduced-budget settings use prefixes of the same full CMA-ES trajectory.
  }
  \label{fig:supp_search_budget}
\end{figure}

\Cref{tab:supp_search_budget,fig:supp_search_budget} show that
target-averaged Dice improves as more search computation is available,
increasing from 76.2 to 77.9 on BraTS and from 68.7 to 71.4 on MAMA-MIA.
Individual targets show small non-monotonic fluctuations, while the aggregate
trend indicates a clear accuracy--search-cost trade-off.

\subsection{Criterion-Ranking Correlation at $B=150$}
\label{app:rank_correlation}

To examine overlap among the eight segmentation criteria, we compute
pairwise Spearman rank correlations between their source-case rankings
separately for each held-out target. The analysis uses the criterion
rankings generated at the representative source budget $B=150$.

\begin{figure*}[!t]
  \centering
  \begin{minipage}[t]{0.48\textwidth}
    \centering
    \includegraphics[width=\linewidth]
    {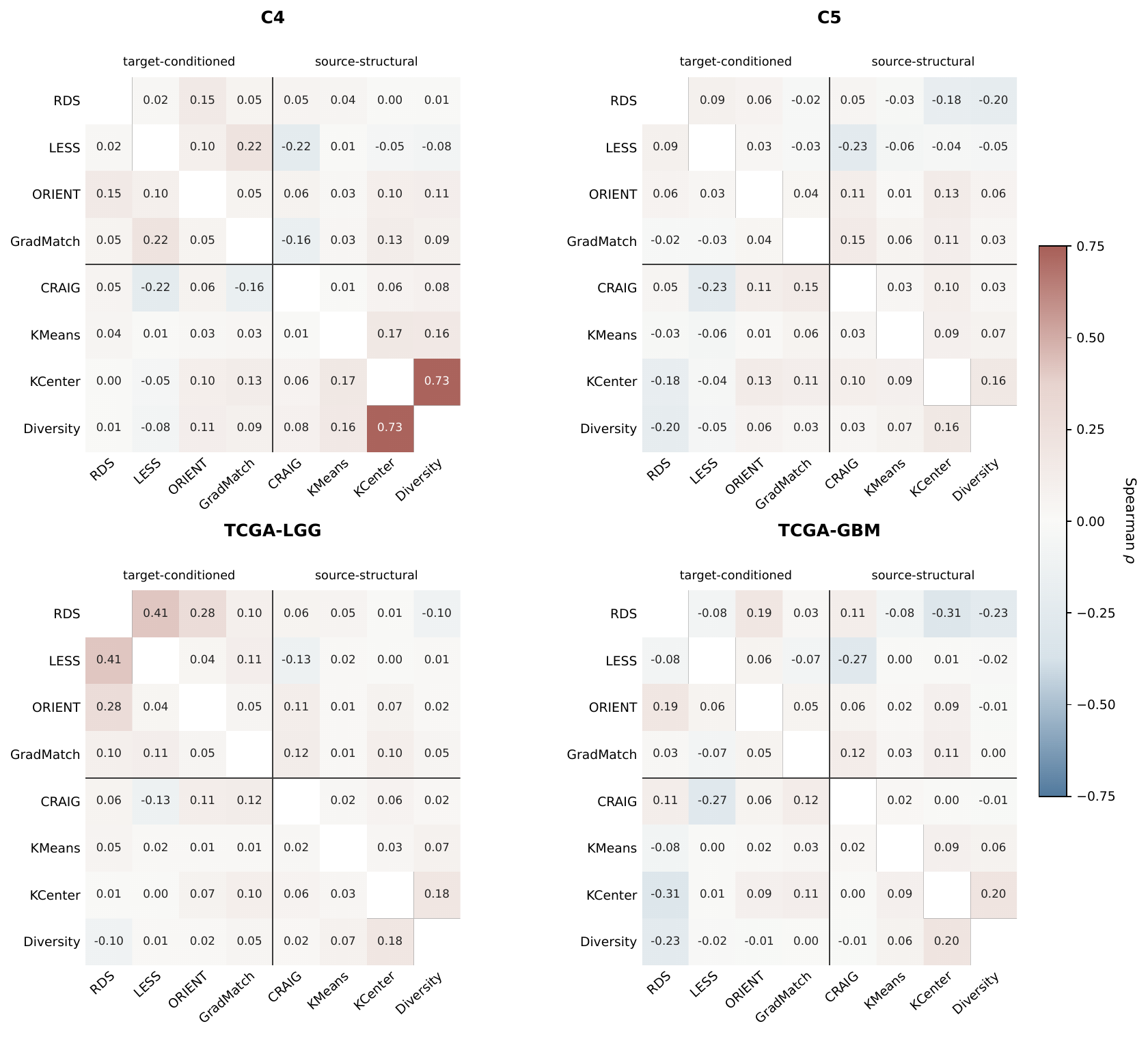}
    \vspace{-1mm}
    \textbf{(a) BraTS}
  \end{minipage}
  \hfill
  \begin{minipage}[t]{0.48\textwidth}
    \centering
    \includegraphics[width=\linewidth]
    {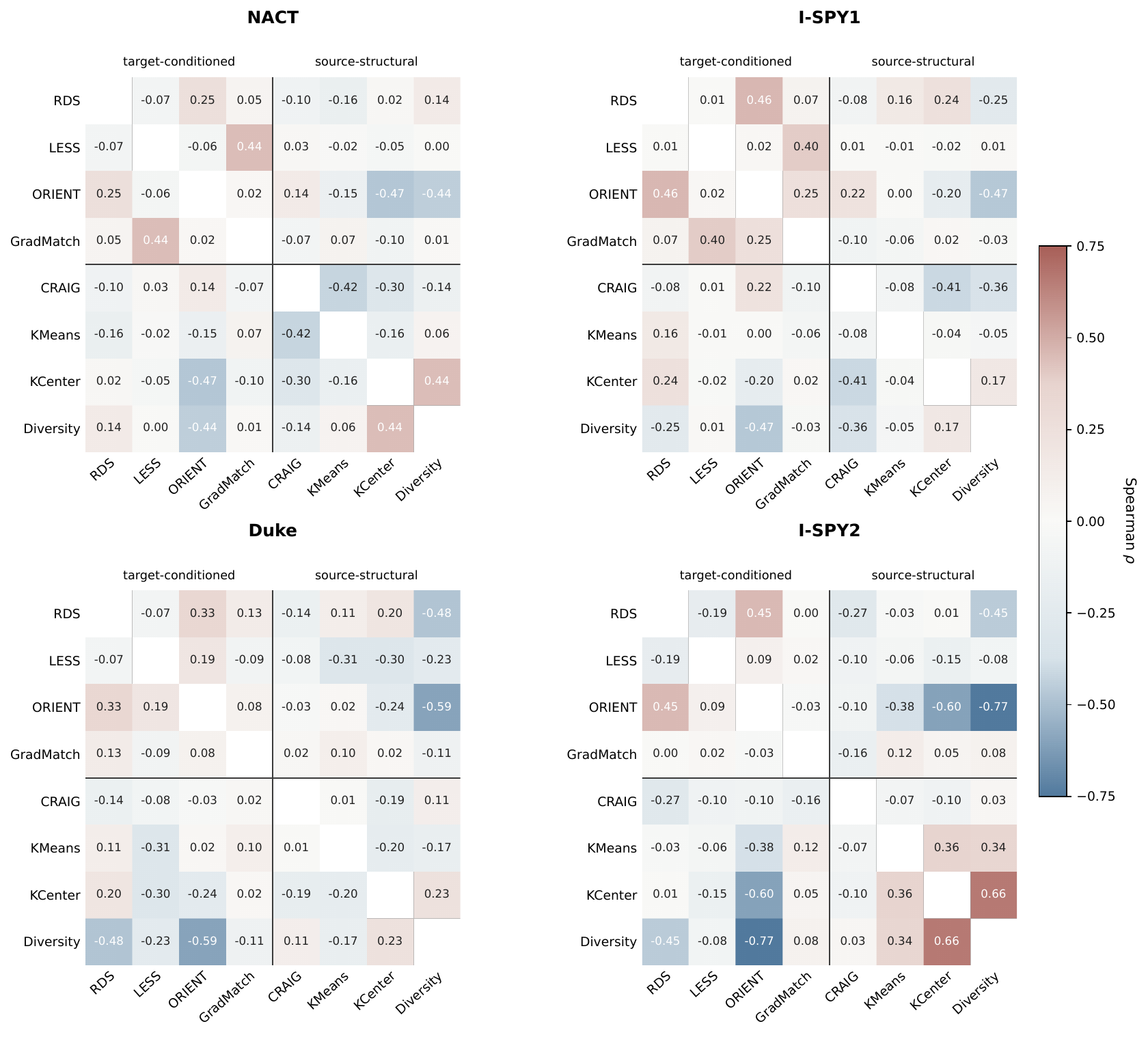}
    \vspace{-1mm}
    \textbf{(b) MAMA-MIA}
  \end{minipage}
  \caption{
  Target-wise Spearman rank correlations among the eight source-utility
  criteria at $B=150$: (a) BraTS and (b) MAMA-MIA.
  }
  \label{fig:supp_rank_correlation}
\end{figure*}

\Cref{fig:supp_rank_correlation} shows that the criterion rankings are not uniformly redundant. On BraTS,
most pairwise correlations are weak-to-moderate, with stronger relationships
appearing only for particular criterion pairs and targets; for example,
KCenter--Diversity reaches $\rho=0.73$ on C4 but is substantially weaker
on the other BraTS targets. MAMA-MIA likewise exhibits structured positive and negative associations, with both their magnitude and pattern varying across targets. Thus, the criteria provide structured but non-identical source rankings,
supporting target-adaptive fusion rather than reliance on a single fixed
criterion.
\subsection{Additional Representative Metrics}
\label{app:additional_metrics}

This subsection reports post-hoc HD95 diagnostics corresponding to the main
$B=150$ tumor-segmentation comparisons. These metrics are not used for TAVO
fusion selection or checkpoint selection; the separate controlled BraTS
nnU-Net HD95 results are reported in Sec.~\ref{app:strong_comparators}.

\begin{table*}[tp]
\centering
\caption{HD95 diagnostics at the representative tumor-segmentation budget
($B=150$). Lower is better.}
\label{tab:supp_hd95_b150}
\small
\setlength{\tabcolsep}{2.4pt}
\renewcommand{\arraystretch}{0.98}
\begin{adjustbox}{max width=\textwidth}
\begin{tabular}{lccccc@{\hspace{1.5em}}ccccc}
\toprule
& \multicolumn{5}{c}{BraTS Macro HD95 \(\downarrow\)}
& \multicolumn{5}{c}{MAMA-MIA Foreground HD95 \(\downarrow\)} \\
\cmidrule(lr){2-6}\cmidrule(lr){7-11}
Method & C4 & C5 & LGG & GBM & Avg.
& NACT & I-SPY1 & Duke & I-SPY2 & Avg. \\
\midrule

Target-only
& \valstd{39.6}{6.2}
& \valstd{25.5}{1.4}
& \valstd{67.3}{2.4}
& \valstd{27.0}{1.2}
& 39.9
& \valstd{38.0}{3.9}
& \valstd{26.7}{4.5}
& \valstd{46.6}{43.1}
& \valstd{24.1}{5.6}
& 33.9 \\

Source-only Full
& \valstd{44.1}{3.8}
& \valstd{28.5}{4.5}
& \valstd{63.0}{3.3}
& \valstd{25.9}{3.9}
& 40.4
& \valstd{23.1}{1.2}
& \valstd{24.6}{2.4}
& \valstd{108.4}{11.2}
& \valstd{30.8}{3.9}
& 46.7 \\

Target + Full Source
& \valstd{36.2}{9.2}
& \valstd{22.1}{3.7}
& \valstd{66.5}{8.1}
& \valstd{22.1}{5.2}
& 36.7
& \valstd{22.7}{0.8}
& \valstd{22.1}{2.9}
& \valstd{92.3}{43.5}
& \valstd{21.2}{8.6}
& 39.6 \\

\midrule

Random
& \valstd{34.7}{8.6}
& \valstd{24.5}{4.1}
& \valstd{70.3}{6.1}
& \valstd{23.4}{2.8}
& 38.2
& \valstd{19.3}{13.2}
& \valstd{18.6}{7.0}
& \valstd{71.3}{36.8}
& \valstd{25.5}{4.9}
& 33.7 \\

\midrule

RDS
& \valstd{35.6}{6.2}
& \valstd{22.5}{2.4}
& \valstd{61.3}{6.4}
& \valstd{21.4}{0.4}
& 35.2
& \valstd{19.5}{3.1}
& \valstd{19.4}{0.9}
& \valstd{52.5}{22.0}
& \valstd{29.1}{3.5}
& 30.2 \\

LESS
& \valstd{36.2}{4.9}
& \valstd{22.5}{2.5}
& \valstd{61.6}{0.2}
& \valstd{21.6}{2.2}
& 35.5
& \valstd{22.6}{4.4}
& \valstd{23.9}{3.2}
& \valstd{41.4}{24.6}
& \valstd{23.8}{3.3}
& 27.9 \\

ORIENT
& \valstd{36.5}{4.3}
& \valstd{13.7}{5.1}
& \valstd{63.3}{1.2}
& \valstd{21.8}{1.0}
& 33.8
& \valstd{25.0}{4.8}
& \valstd{21.9}{0.5}
& \valstd{49.5}{42.6}
& \valstd{25.1}{5.9}
& 30.4 \\

GradMatch
& \valstd{33.1}{6.8}
& \valstd{22.8}{1.6}
& \valstd{64.6}{7.1}
& \valstd{19.2}{4.8}
& 34.9
& \valstd{16.9}{1.7}
& \valstd{22.1}{2.2}
& \valstd{51.9}{31.9}
& \valstd{22.6}{6.6}
& 28.4 \\

CRAIG
& \valstd{36.0}{4.4}
& \valstd{25.5}{1.5}
& \valstd{65.8}{6.5}
& \valstd{21.3}{2.9}
& 37.2
& \valstd{22.0}{19.8}
& \valstd{23.3}{1.8}
& \valstd{81.7}{25.6}
& \valstd{25.4}{5.2}
& 38.1 \\

KMeans
& \valstd{39.6}{7.7}
& \valstd{19.3}{0.0}
& \valstd{74.4}{8.7}
& \valstd{21.1}{1.9}
& 38.6
& \valstd{25.9}{1.5}
& \valstd{20.6}{2.3}
& \valstd{95.0}{13.2}
& \valstd{23.3}{6.8}
& 41.2 \\

KCenter
& \valstd{38.7}{14.2}
& \valstd{17.3}{2.1}
& \valstd{67.2}{2.7}
& \valstd{20.5}{3.2}
& 35.9
& \valstd{24.1}{2.0}
& \valstd{22.4}{3.2}
& \valstd{70.5}{14.0}
& \valstd{22.0}{9.6}
& 34.8 \\

Diversity
& \valstd{36.6}{1.5}
& \valstd{25.3}{3.6}
& \valstd{68.4}{4.0}
& \valstd{23.2}{3.3}
& 38.4
& \valstd{22.4}{11.2}
& \valstd{26.0}{2.6}
& \valstd{65.6}{39.2}
& \valstd{26.7}{4.5}
& 35.2 \\

\midrule

DANN
& \valstd{37.5}{7.7}
& \valstd{19.7}{2.8}
& \valstd{68.2}{4.5}
& \valstd{19.4}{3.9}
& 36.2
& \valstd{25.4}{7.1}
& \valstd{29.7}{7.8}
& \valstd{58.2}{14.5}
& \valstd{25.4}{7.9}
& 34.7 \\

MMD
& \valstd{38.9}{9.6}
& \valstd{19.7}{3.3}
& \valstd{70.1}{1.3}
& \valstd{20.6}{1.3}
& 37.3
& \valstd{28.9}{9.6}
& \valstd{21.7}{3.3}
& \valstd{67.1}{34.3}
& \valstd{29.6}{5.9}
& 36.8 \\

ADVENT
& \valstd{30.5}{4.7}
& \valstd{23.8}{0.4}
& \valstd{64.8}{1.8}
& \valstd{20.0}{0.4}
& 34.8
& \valstd{23.5}{4.7}
& \valstd{27.8}{3.4}
& \valstd{54.8}{21.8}
& \valstd{28.0}{5.4}
& 33.5 \\

SE-ASA
& \valstd{33.2}{8.1}
& \valstd{22.9}{5.4}
& \valstd{70.0}{2.7}
& \valstd{25.5}{2.7}
& 37.9
& \valstd{26.2}{8.1}
& \valstd{23.9}{1.4}
& \valstd{50.0}{22.7}
& \valstd{31.5}{5.1}
& 32.9 \\

\midrule

TAVO
& \valstd{32.8}{9.4}
& \valstd{19.6}{2.7}
& \valstd{59.9}{6.9}
& \valstd{21.0}{1.3}
& 33.3
& \valstd{24.2}{3.9}
& \valstd{19.4}{2.4}
& \valstd{42.6}{27.4}
& \valstd{25.6}{3.4}
& 28.0 \\

\bottomrule
\end{tabular}
\end{adjustbox}
\end{table*}
\section{Reproducibility Notes}
\label{app:reproducibility}

This appendix records the practical details needed to check and reproduce the
reported experiments. The accompanying code package contains the scripts used
for split construction, source selection, TAVO score fusion, final-training
command generation, result aggregation, and basic integrity checks. It does not
redistribute raw datasets, preprocessed medical caches, trained checkpoints,
logs, or generated experimental outputs.

\subsection{Environment}
\label{app:environment}

\Cref{tab:supp_environment} reports the software and hardware environment used
for representative measured costs. The provided code uses relative paths and
expects dataset roots, generated split files, score files, and output
directories to be supplied at runtime.

\begin{table}[ht]
\centering
\caption{Software and hardware environment used for representative measured
costs and experiment execution.}
\label{tab:supp_environment}
\small
\setlength{\tabcolsep}{4pt}
\renewcommand{\arraystretch}{1.10}
\begin{tabular}{ll}
\toprule
Item & Value \\
\midrule
GPU for measured medical costs & NVIDIA A40 \\
Operating system & Rocky Linux 8.10 \\
CUDA / cuDNN & CUDA 12.8 / cuDNN 9.10.2 \\
Python & 3.10.16 \\
Core Python dependencies & NumPy, PyYAML, Pillow \\
Medical dependencies & nibabel, SciPy \\
Training dependencies & PyTorch, torchvision \\
nnU-Net & nnU-Net v2 backend \\
EfficientViT & EfficientViT-Seg-L1 backend \\
Split seed & 2026 \\
Final-training seeds & 0, 1, 2 \\
\bottomrule
\end{tabular}
\end{table}

\subsection{Run Records}
\label{app:run_records}

\paragraph{Randomness and run identity.}
Dataset split manifests were constructed once using split seed 2026 and remain
fixed across methods, source budgets, and final-training seeds. Final reported
models use training seeds 0, 1, and 2. The CMA-ES sampling seed is independent
of the neural-network training seeds, while the coarse/refinement seed sets and
search configuration are fixed as described in Sec.~A.3.

Each reported result is determined by the dataset, held-out target, source
budget, method, split seed, final-training seed, selected-source list,
configuration file, checkpoint-selection rule, and seed-level target-test
metric export. For the clinical datasets, provided selected-source lists use
hashed case identifiers rather than raw clinical identifiers. Reported mean$\pm$standard deviation values are computed from the corresponding
per-seed target-test exports.

\subsection{Dataset Access}
\label{app:dataset_access}

BraTS 2021, MAMA-MIA, and OfficeHome are used under their respective dataset
licenses and access terms. The code package does not include raw medical
images, clinical annotations, private metadata, preprocessed caches, trained
model weights, logs, or generated experimental outputs. Users who wish to
rerun the experiments should obtain the datasets through the official channels
and point the provided configuration files to their local dataset roots.
\Cref{tab:supp_dataset_access} summarizes the dataset access requirements.

\begin{table}[ht]
\centering
\caption{Dataset access requirements for reproducing the experiments.}
\label{tab:supp_dataset_access}
\small
\setlength{\tabcolsep}{3.0pt}
\renewcommand{\arraystretch}{1.12}
\begin{tabularx}{\linewidth}{
l
>{\raggedright\arraybackslash}p{0.27\linewidth}
>{\raggedright\arraybackslash}X}
\toprule
Dataset & Data used & Access note \\
\midrule
BraTS 2021 &
Images, tumor labels, center/domain mapping &
Available through the official BraTS 2021 challenge/data access process;
raw images and labels are not redistributed. \\

MAMA-MIA &
DCE-MRI images, tumor labels, cohort/domain mapping &
Available under the official MAMA-MIA dataset access terms; raw images and
clinical annotations are not redistributed. \\

OfficeHome &
Images and domain labels &
Public OfficeHome dataset; experiments use the Art, Clipart, Product, and
Real World domains. \\
\bottomrule
\end{tabularx}
\end{table}
\FloatBarrier
\clearpage
\endgroup

\end{document}